\pdfoutput=1

\documentclass[11pt]{article}

\usepackage[preprint]{coling}

\usepackage{times}
\usepackage{latexsym}

\usepackage[T1]{fontenc}

\usepackage[utf8]{inputenc}
\usepackage[main=english,russian,ukrainian,spanish]{babel}

\usepackage{microtype}

\usepackage{inconsolata}

\usepackage{graphicx}
\usepackage{array}
\usepackage{hyperref}
\usepackage{natbib}
\usepackage{booktabs}
\usepackage{color, colortbl}
\usepackage{enumitem}
\usepackage{caption}
\usepackage{adjustbox}
\usepackage{amssymb}
\usepackage{utfsym}
\usepackage{mathtools}
\usepackage{amsmath}
\usepackage{inconsolata}
\usepackage{subcaption}
\usepackage{enumitem}
\usepackage{xcolor}
\usepackage[most]{tcolorbox}
\usepackage{textcomp}

\definecolor{Lightgreen}{RGB}{165, 224, 209}
\definecolor{Gray}{gray}{0.9}

\tcbset{
  colback=black!5!white,        
  colframe=gray!10!black,      
  width=\linewidth,       
  boxrule=0.4mm,          
  arc=2mm,                
  outer arc=2mm,          
  boxsep=2.5mm,             
  title=Prompt Instructions
}

\title{Multilingual and Explainable Text Detoxification \\ with Parallel Corpora}

\author{
\textbf{Daryna Dementieva}\textsuperscript{1}, \textbf{Nikolay Babakov}\textsuperscript{2}, \\ \textbf{Amit Ronen}\textsuperscript{3}, \textbf{Abinew Ali Ayele}\textsuperscript{4,5}, \textbf{Naquee Rizwan}\textsuperscript{6}, \textbf{Florian Schneider}\textsuperscript{4}, \\ \textbf{Xintong Wang}\textsuperscript{4}, \textbf{Seid Muhie Yimam}\textsuperscript{4}, \textbf{Daniil Moskovskiy}\textsuperscript{8,9}, \textbf{Elisei Stakovskii}\textsuperscript{10}, \\ \textbf{Eran Kaufman}\textsuperscript{3}, \textbf{Ashraf Elnagar}\textsuperscript{7}, \textbf{Animesh Mukherjee}\textsuperscript{6}, \textbf{Alexander Panchenko}\textsuperscript{8,9} \\
\footnotesize{
\textsuperscript{1}Technical University of Munich, \textsuperscript{2}Universidade de Santiago de Compostela, 
\textsuperscript{3}Shenkar College,} \\
\footnotesize{
\textsuperscript{4}University of Hamburg, 
\textsuperscript{5}Bahir Dar University,
\textsuperscript{6}IIT Kharagpur, \textsuperscript{7}University of Sharjah,}\\
\footnotesize{
\textsuperscript{8}Skoltech, \textsuperscript{9}AIRI, \textsuperscript{10}University of North Carolina at Chapel Hill} \\ 
\href{mailto:daryna.dementieva@tum.de}{\texttt{\small daryna.dementieva@tum.de}},
\href{mailto:a.panchenko@skol.tech}{\texttt{\small a.panchenko@skol.tech}}
}

\begin{document}
\maketitle
\begin{abstract}
Even with various regulations in place across countries and social media platforms~\cite{india_it_rules_2021,eu_digital_services_act_2022}, digital abusive speech remains a significant issue. One potential approach to address this challenge is automatic text detoxification, a text style transfer (TST) approach that transforms toxic language into a more neutral or non-toxic form. To date, the availability of parallel corpora for the text detoxification task~\cite{logacheva-etal-2022-paradetox,DBLP:conf/coling/AtwellHA22,dementieva2024multiparadetox} has proven to be crucial for state-of-the-art approaches. With this work, we extend parallel text detoxification corpus to new languages---German, Chinese, Arabic, Hindi, and Amharic---testing in the extensive multilingual setup TST baselines. Next, we conduct the first of its kind an automated, explainable analysis of the descriptive features of both toxic and non-toxic sentences, diving deeply into the nuances, similarities, and differences of toxicity and detoxification across 9 languages. Finally, based on the obtained insights, we experiment with a novel text detoxification method inspired by the Chain-of-Thoughts reasoning approach, enhancing the prompting process through clustering on relevant descriptive attributes. \\
\textcolor{red}{\textit{Warning: This paper contains offensive texts that only serve as illustrative examples.}}
\end{abstract}

\section{Introduction}

\begin{figure}[ht!]
    \centering
    \includegraphics[width=1\linewidth]{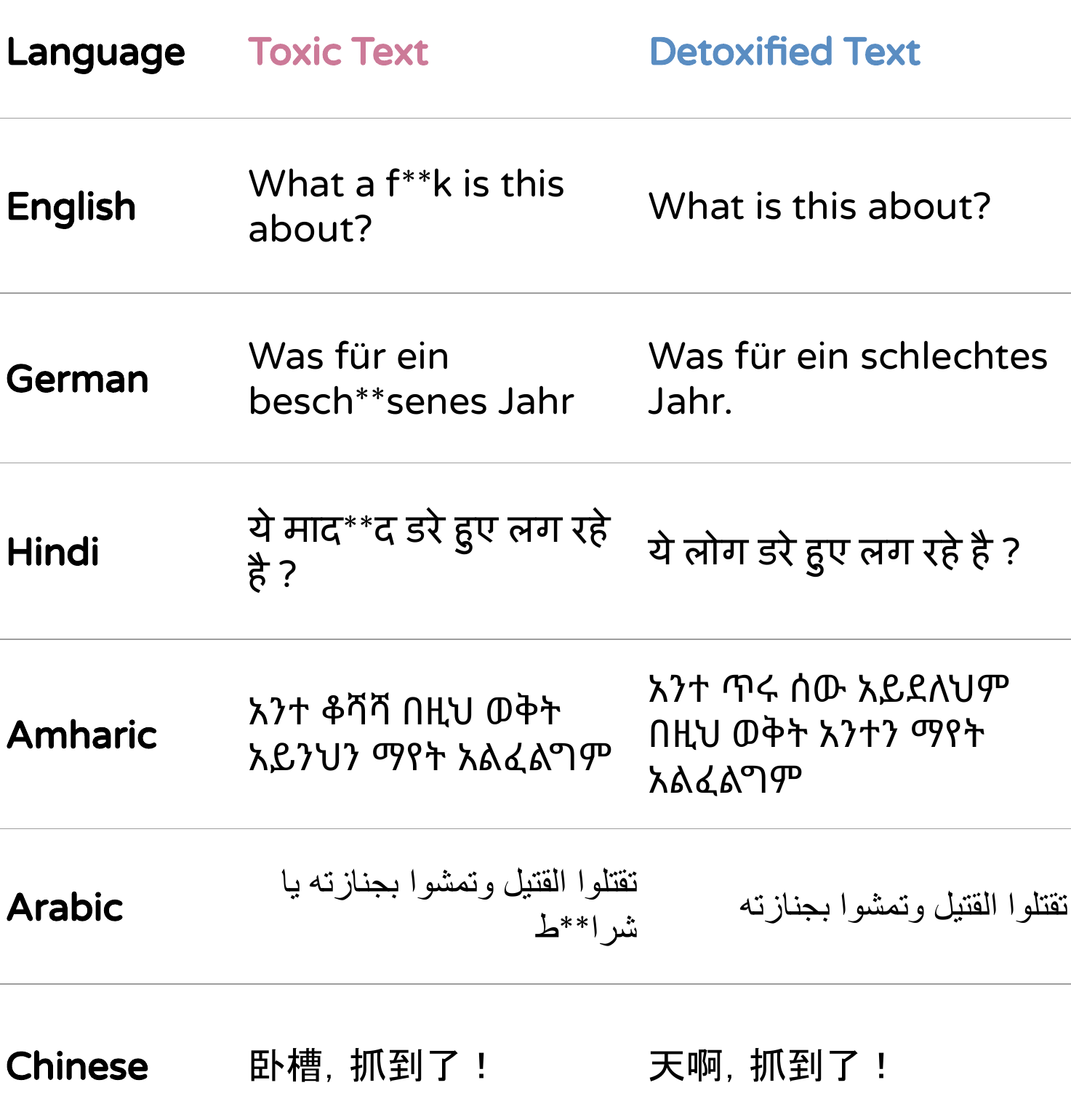}
    \caption{Examples of the desired texts detoxification for English and new languages: German, Chinese, Arabic, Hindi, and Amharic.}
    \label{fig:intro}
\end{figure}

The issue of managing toxic speech remains a crucial aspect of human communication and \textbf{digital violence} prevention~\cite{DBLP:journals/corr/abs-2001-01818}, including the mitigation of toxic responses generated by Large Language Models (LLMs)~\cite{DBLP:journals/corr/abs-2312-02003}. The typical approach to dealing with abusive speech on social platforms involves message blocking~\cite{cobbe2021algorithmic}. To address this, numerous toxic and hate speech detection models have been developed for different languages, i.e. English~\cite{DBLP:conf/aaai/MathewSYBG021}, Spanish~\cite{DBLP:journals/access/MoleroPRP23}, Amharic~\cite{ayele-etal-2023-exploring}, Code-Mixed Hindi~\cite{bohra-etal-2018-dataset}, and many others~\cite{DBLP:journals/corr/abs-2401-05060}. However, the recent research indicates a necessity for more proactive moderation of abusive speech~\cite{kulenovic2023should}. One such approach is \textbf{text detoxification}.

\begin{figure*}[h!]
    \centering
    \includegraphics[width=\textwidth]{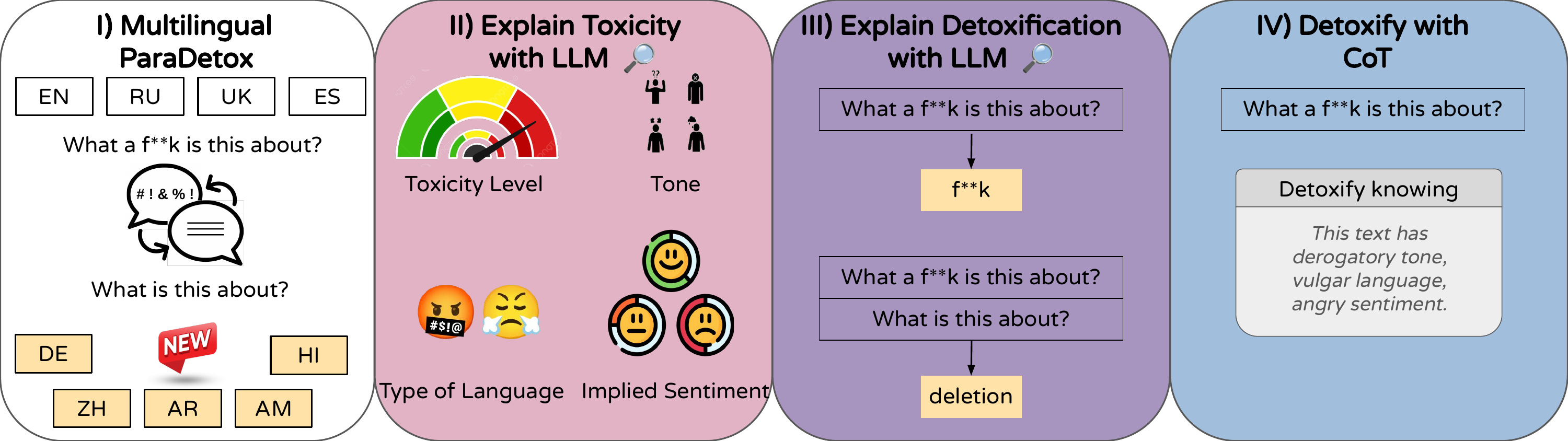}
    \caption{In this work, we extend parallel text detoxification data to new languages as well as provide explainability analysis of toxicity and detoxification attributes across all languages. This information helps to improve Chain-of-Thoughts reasoning for automatic text detoxification with LLMs.}
    \label{fig:textdetox-logo}
\end{figure*}

Within the baselines approaches for automatic text detoxification, multiple unsupervised baselines were created based on ideas of Delete-Retrieve-Generate~\cite{DBLP:conf/naacl/LiJHL18}, latent style spaces disentanglement~\cite{nogueira-dos-santos-etal-2018-fighting}, or conditional generation with Masked Language Modeling~\cite{dale-etal-2021-text}. However, the latest state-of-the-art outcomes, particularly in English, were attained when parallel data and fine-tuning with text-to-text generation models were employed as in ParaDetox~\cite{logacheva-etal-2022-paradetox} or APPDIA~\cite{DBLP:conf/coling/AtwellHA22}. Then, several works were conducted to explore the potential of multilingual and cross-lingual text detoxification~\cite{moskovskiy-etal-2022-exploring,dementieva2024multiparadetox}. With this work, we extend the parallel text detoxification corpora to even more languages. Also, we are the first to conduct a comprehensive analysis of the full parallel multilingual corpus, uncovering unique traits and commonalities in how toxicity manifests across different languages and the ways to rephrase them.
Thus, our contributions are the following (see Figure~\ref{fig:textdetox-logo}):
\begin{itemize}
    \item We extend parallel text detoxification data to new languages---German, Chinese, Arabic, Hindi, and Amharic---thoroughly reporting each annotation process (Figure~\ref{fig:textdetox-logo}, I);
    \item We perform the first-of-its-kind study on explainability of parallel detoxification data thoroughly examining toxicity (Figure~\ref{fig:textdetox-logo}, II) and detoxification attributes (Figure~\ref{fig:textdetox-logo}, III) across 9 languages;
    \item Finally, we benchmark text detoxification baselines across a comprehensive multilingual dataset, incorporating a novel Chain-of-Thoughts prompting approach for detoxification with LLMs (Figure~\ref{fig:textdetox-logo}, IV).
\end{itemize}

All data, code, and analysis results are publicly accessible online.\footnote{\href{https://huggingface.co/textdetox}{https://huggingface.co/textdetox}}$^{,}$\footnote{\href{https://github.com/textdetox/multilingual_explainable_paradetox}{https://github.com/textdetox/\\multilingual\_explainable\_paradetox}}$^{,}$\footnote{The data introduced in this work served as the foundation for the TextDetox CLEF-2024 Shared Task~\cite{dementieva2024overview}.}



\section{Related Work}

\paragraph{Modern Text Style Transfer} Text style transfer (TST) methods can generally be categorized into unsupervised and supervised approaches~\cite{jin-etal-2022-deep}. Typically, when a text classification corpus for a specific domain is available, unsupervised methods are employed. For instance, condBERT and ParaGedi were introduced for controllable masked language modeling in~\cite{dale-etal-2021-text}, with MaRCo further enhancing these methods by incorporating multiple experts~\cite{hallinan-etal-2023-detoxifying}. Additionally, diffusion models have been explored for controllable text generation, particularly for text detoxification~\cite{floto-etal-2023-diffudetox,DBLP:conf/aaai/HorvitzPCYM24}. Large Language Models (LLMs) have also shown promising results across various NLP tasks, including paraphrasing, leading to their application in different TST tasks~\cite{DBLP:journals/corr/abs-2406-05885}, and specifically in text detoxification through the CoTex pipeline~\cite{zhang-etal-2024-distilling}. However, the availability of parallel training corpora has been shown to significantly enhance the performance of TST methods, often surpassing LLMs, which can be prone to hallucination. Such parallel corpora, though, are limited to specific tasks, including Bible historical styles~\cite{carlson2018evaluating}, GYAFC for formality~\cite{rao-tetreault-2018-dear}, and APPDIA~\cite{DBLP:conf/coling/AtwellHA22} and ParaDetox~\cite{logacheva-etal-2022-paradetox} for detoxification.

\paragraph{Multilingual Text Style Transfer} To date, several studies have explored text style transfer across various languages, extending beyond just English. For instance, sentiment transfer has been developed for Bangla~\cite{mukherjee-etal-2023-low} and other Indian languages~\cite{DBLP:journals/corr/abs-2405-20805}. In terms of formality, the English-focused GYAFC dataset was expanded to the X-FORMAL dataset~\cite{briakou-etal-2021-ola}, which includes Brazilian Portuguese, French, and Italian. More recently, formality style transfer has been examined for Japanese~\cite{ung2023formality}. Detoxification techniques have been applied to English~\cite{logacheva-etal-2022-paradetox}, then Russian, Ukrainian, and Spanish~\cite{dementieva2024multiparadetox}. However, these studies still primarily focus on European languages, leaving many other regions of the world unexplored.
%

\paragraph{Explainable Abusive Speech Mitigation} 
To build trustworthy systems for mitigating different kinds of abusive speech, the aspect of explainablility has gained increasing attention recently~\cite{DBLP:journals/jocss/GonganeMA24}. One of the first work in this area~\cite{DBLP:conf/aaai/MathewSYBG021} introduced the HateXplaine dataset, where annotators not only labeled the data but also provided the rationale behind their classifications. Following this, explainable AI frameworks like SHAP~\cite{DBLP:conf/nips/LundbergL17} and LIME~\cite{DBLP:conf/kdd/Ribeiro0G16} have been applied to various text classification tasks, including hate and toxic speech~\cite{mosca-etal-2023-ifan,imbwaga2024explainable}. For toxic language specifically, the ToXCL framework~\cite{hoang-etal-2024-toxcl} was developed to fine-tune multiple models addressing different aspects of toxic speech detection. Additionally, recent advancements in LLMs have been leveraged for both text style transfer and generating corresponding explanations in the context of text detoxification~\cite{DBLP:journals/corr/abs-2402-15951}.

\section{New ParaDetox Annotation}
We manually collected new data following the main quality criteria~\cite{logacheva-etal-2022-paradetox}: (i)~new paraphrases should be non-toxic; (ii)~maximal content preservation; (iii)~fluency on par with the original text. These data cover \textbf{five languages}---German, Hindi, Amharic, Arabic, and Chinese---chosen based on the native languages of the authors. Annotation and quality control were conducted either by the authors themselves or by hired assistants fluent in the respective languages.

\paragraph{Definition of Toxicity} We adopt the definition introduced by~\citet{dementieva2024multiparadetox} only addressing \textbf{vulgar or profane language}~\cite{DBLP:journals/corr/abs-2207-04672,logacheva-etal-2022-paradetox} while the overall message can be either toxic or neutral, but it should not involve deep insults or hate towards individuals or groups of people.

\begingroup
\renewcommand{\arraystretch}{1.15}
\begin{table}[ht!]
    \centering
    \footnotesize
    \begin{tabular}{p{1.2cm}|c|c|c|c}
    \toprule
    \shortstack{\textbf{Language} \\ \ } & \textbf{\shortstack{\# \\ Annot.}} & \textbf{\shortstack{\# Annot. \\ Per Sent.}}& \textbf{\shortstack{\# Toxic \\ Sent.}} & \textbf{\shortstack{\# \\ Detoxified}} \\
    \midrule
        German & 2 & 1 & 3\,521 & 1\,103 \\
        Hindi & 2 & 1 & 2\,328 & 1\,007 \\
        Amharic & 2 & 1 & 2\,995 & 1\,000\\
        Arabic & 3 & 3 & 2100 & 1181 \\
        Chinese & 3 & 1 & 1\,380 & 1\,000\\
    \bottomrule
    \end{tabular}
    \caption{Summary of annotators and detoxifiable sentences statistics per language.}
    \label{tab:annotation_statistics}
\end{table}
\endgroup

\paragraph{Data Preprocessing} For all languages, we maintain the length of samples as sentences of around 5-20 tokens. Also, if a text sample is from a social network, we anonymize or fully eliminate any mentioning of usernames and links.

\paragraph{Annotation Guidelines} Firstly, we organized a joint meeting with all language stakeholders to present the base annotation guidelines in English. If needed, each language stakeholder adapted these guidelines to their own language and cultural context; otherwise, they reused the English version. All final annotation guidelines are made publicly available online.\footnote{\href{https://github.com/textdetox/multilingual_explainable_paradetox/tree/main/paradetox_collection}{https://github.com/textdetox/multilingual\_explainable\_\\paradetox/tree/main/paradetox\_collection}}

\paragraph{Annotators Compensations} Compensation varied according to each university’s and country’s regulations. For German and Chinese, annotators were employed in Germany at a rate of \texteuro20 per hour (\texteuro7.65 above the minimum wage). For Amharic, the annotators were hired from Ethiopia \$5.8 per hour which is better than an M.Sc holder salary in the country. For Arabic and Hindi data, the annotators were existing lab or research projects employees receiving standard academic salaries.

\paragraph{Annotators Well-Being} The annotation process took about four months providing flexible schedules and regular check-ins. Language stakeholders and task experts met weekly to discuss issues; daily stand-ups with annotators ensured supportive progress. Annotators could pause at any time without meeting daily quotas. Their expertise suited the project's needs, and limitations emerged.


\subsection{German}
\label{sec:german}

German ParaDetox was collected with several annotators with manual quality verification:

\subsubsection{Input Data Preparation}
\label{sec:german_data}

The German language source data is based on three datasets containing toxic, offensive, or hate speech comments on social media about primarily political events in Germany or the US.
For the two datasets from the GermEval 2018~\citep{wiegand2018germeval} and GermEval 2021~\citep{risch2021germeval} shared tasks, we used data from both the test and the train split.
For the GermEval 2018 data, we only used samples labeled with the coarse class ``\textit{OFFENSE}'' whereas for the GermEval 2021 data we only used samples annotated with the ``\textit{Sub1\_Toxic}'' class.
The third dataset~\citep{ross2016hatespeech} was filtered so only samples were kept where both expert annotators classified the samples as hate speech.
The data from the three datasets was merged and deduplicated via exact string matching. As a result, 3\,521 toxic were selected as candidates from which 1\,103 were possible to detoxify.

\subsubsection{Annotation Process}
\label{sec:german_annotations}
To create the final parallel detoxified German dataset, we hired two native German annotators.
Annotator A is a female born in $1994$ who holds a Master of Arts degree in Social Sciences, and Annotator B is a male born in $1992$ who holds a Master of Science degree in Computer Science.
The data was distributed so that each sample was transcribed by only one of the annotators.
%

%
%

\subsection{Hindi}
\label{sec:hindi}

Hindi dataset was collected manually by native-speakers gaining data from multiple sources:


\subsubsection{Input Data Preparation}

We used the HASOC dataset created at FIRE 2019~\cite{10.1145/3368567.3368584} as source for Hindi language. Contents in this dataset are relevant within Indian subcontinent which are collected from various social media platforms prevalent in India.
For curation, posts containing \textit{OFFENSIVE} and \textit{PROFANE} contents in train and test splits were used. 1\,455 \textit{PROFANE} posts (1\,237 train + 218 test) and 873 \textit{OFFENSIVE} posts (676 train + 197 test) were chosen to prepare detoxifiable toxic data for our task.
%
On a total of 2\,328 samples, we first performed deduplication via exact string matching.


\subsubsection{Annotation Process}

\paragraph{Annotation Setup}

Out of 2328 samples, 1007 samples were marked as detoxifiable. 
Annotators were guided to re-write toxic pairs in a non-toxic manner, keeping the meaning of the original post unchanged.
\paragraph{Annotators} One male NLP researcher working in the field of hate/toxic speech and another female student enrolled in Bachelor's Degree and having experience in Machine Learning, were employed to carry out the annotation. Both annotators are Indian, native Hindi speakers and are well versed with the topicality covered in the dataset. Each sentence was assigned to a single annotator. Afterwards, the data were cross-verified by a language stakeholder and domain experts.


\subsection{Amharic}
\label{sec:amharic}

We compiled new Amharic ParaDetox datasets with the following annotation details:

\subsubsection{Input Data Preparation}


The input toxicity data is entirely sourced from the two previous studies, namely \cite{ayele-etal-2023-exploring} and \cite{ayele20225js}. We extracted a subset of these datasets labeled as \emph{offensive}.


\subsubsection{Annotation Process}
\paragraph{Annotation Setup} We customized the Potato-POrtable Text Annotation TOol\footnote{\href{https://github.com/davidjurgens/potato}{https://github.com/davidjurgens/potato}} and utilized it for the annotation of Amharic ParaDetox dataset. Annotators were provided annotation guidelines, took hands-on practical training, completed independent training tasks before the main annotation task. 

We began with a pilot annotation of 125 items by three native Amharic speakers and reviewed the quality in a group meeting with experts to clarify the task. Next, we annotated 2\,995 tweets, each by a single annotator. Each tweet was classified as either detoxifiable or non-detoxifiable. Detoxifiable tweets were then rewritten in a detoxified manner.



\paragraph{Annotators} Two annotators (one male and one female) were evolved in the main annotation, where both of them are university lecturers and have basic knowledge of NLP tasks.

\begingroup
\renewcommand{\arraystretch}{1.15}
\begin{table*}[h!]
    \centering
    \footnotesize
    \begin{tabular}{l|c|c|c|c}
    \toprule
       \textbf{Language}  & \shortstack{\bf Source of \\ \bf Toxic Samples} & \textbf{Annotation Process}  & \textbf{Train} & \textbf{Test} \\
       \midrule
        English & \cite{jigsaw} & Crowdsourcing & $400$ & $600$ \\
        Russian & \cite{tox_ru_comments_1,tox_ru_comments_2} & CrowdSourcing & $400$ & $600$ \\
        Ukrainian & \cite{bobrovnyk2019automated} & Crowdsourcing & $400$ & $600$ \\
        \shortstack{Spanish \\ \ } & \shortstack{\cite{DBLP:journals/sensors/Pereira-Kohatsu19,taule2024newscom} \\ \cite{perez-etal-2022-robertuito}} & \shortstack{Crowdsourcing \\ \ } & \shortstack{$400$ \\ \ } & \shortstack{$600$ \\ \ } \\
        \midrule
        \shortstack{German \\ \ } & \shortstack{\cite{wiegand2018germeval,risch2021germeval} \\ \cite{ross2016hatespeech}} & \shortstack{Manual \\ \ } & \shortstack{$400$ \\ \ } & \shortstack{$600$ \\ \ } \\
        Hindi & \cite{10.1145/3368567.3368584} & Manual & $400$ & $600$ \\
        Amharic & \cite{ayele-etal-2023-exploring,ayele20225js} & Manual & $400$ & $600$ \\
        \shortstack{Arabic \\ \ } & \shortstack{\cite{mulki2019hsab,haddad2019t} \\ \cite{mubarak2020overview,mulki-ghanem-2021-mi}} & \shortstack{Manual \\ \ } & \shortstack{$400$ \\ \ } & \shortstack{$600$ \\ \ } \\
        Chinese & \cite{lu-etal-2023-facilitating} & Manual & $400$ & $600$ \\
    \bottomrule
    \end{tabular}
    \caption{All currently available ParaDetox datasets from previous work~\cite{logacheva-etal-2022-paradetox,dementieva2024multiparadetox} and the new ones. The human detoxified references were collected either via crowdsourcing or by hired native speakers. In this work, 1\,000 samples per language were selected to perform analysis and experiments.}
    \label{tab:final_paradetox_dataset}
\end{table*}
\endgroup




\subsection{Arabic}
\label{sec:arabic}

Here are details of Arabic ParaDetox collection:

\subsubsection{Input Data Preparation}
The Arabic ParaDetox dataset was created by combining parts of several existing datasets along with the Arabic-translated version of the Jigsaw dataset~\cite{jigsaw}. It includes the Levantine Twitter Dataset for Hate Speech and Abusive Language (L-HSAB)~\cite{mulki2019hsab}, which focuses on Levantine dialects, and the Tunisian Hate and Abusive Speech (T-HSAB) dataset~\cite{haddad2019t}, which targets Tunisian dialects. It also incorporates the OSACT dataset~\cite{mubarak2020overview} and the Arabic Levantine Twitter Dataset for Misogynistic Language (LeT-Mi)~\cite{mulki-ghanem-2021-mi}, which specifically addresses gender-based abuse. These resources combine to form the Arabic ParaDetox dataset, aimed at aiding the development of toxicity classifiers capable of handling Arabic content across various dialects and contexts. As a result, 2100 sentences were selected as candidates with 1181 were possible to detoxify.

\subsubsection{Annotation Process}

\paragraph{Annotators}
Detoxification was performed by three PhD-level annotators (two male, one female), all native Arabic speakers with strong computational linguistics backgrounds. Each text sample was transcribed by two annotators, and majority voting determined whether a sentence could be detoxified and if the resulting detoxification was appropriate.

\subsection{Chinese}
\label{sec:chinese}

We collected new Chinese ParaDetox datasets with the following annotation details:

\subsubsection{Input Data Preparation}

\paragraph{Input Toxicity Data} The Chinese ParaDetox dataset is derived from TOXICN \cite{lu-etal-2023-facilitating}, a recently released Chinese toxic language dataset. TOXICN was compiled from social media platforms and comprises 12\,011 comments addressing several sensitive topics, including gender, race, region, and LGBTQ issues. From this dataset, we extracted a subset based on multiple criteria: the number of toxic words, the ratio of toxic words in the comments, the length of comments, and the toxic scores of comments.

\paragraph{Input Preprocessing} We set thresholds for the criteria: the number of toxic words ranged from 1 to 5 (checked by the predefined keywords list), the ratio of toxic words in comments was less than 0.5, and the length of comments ranged from 3 to 50 words, ensuring suitability for annotators to rewrite them.
%
Subsequently, we employed a pre-trained toxic classifier \cite{lu-etal-2023-facilitating} to compute the toxic scores of the selected comments, using a threshold score of $0.978$ to filter the candidates. Ultimately, we collected 1\,149 samples from the training set and 231 samples from the test set, resulting in a total of 1\,380 samples deemed suitable for annotation.

\begin{figure*}[ht!]
    \centering
    \begin{subfigure}[b]{0.49\textwidth}
        \includegraphics[width=1\linewidth]{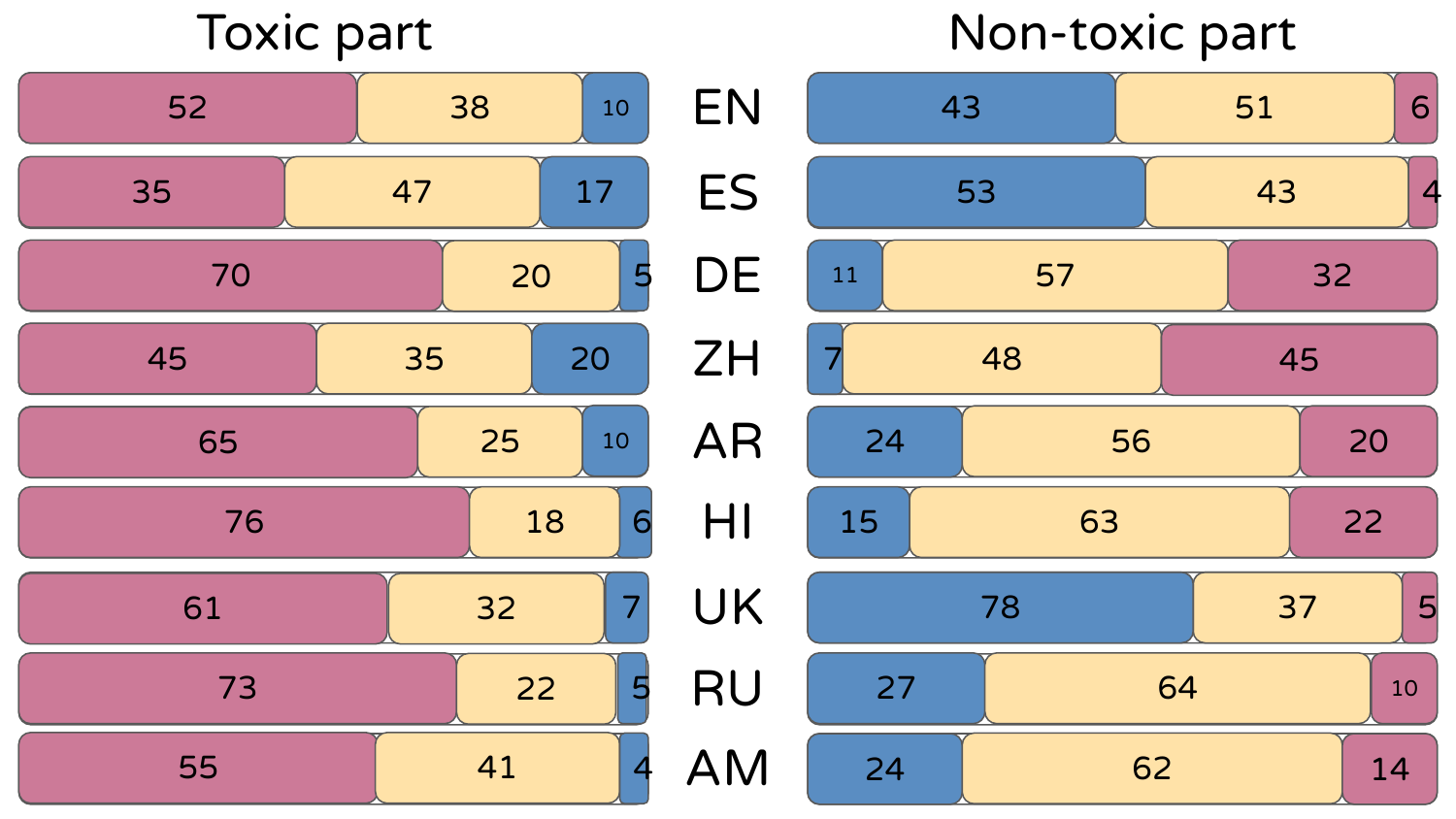}
        \caption{Toxicity Levels}
    \end{subfigure}
    \hfill
    \begin{subfigure}[b]{0.49\textwidth}
        \includegraphics[width=1\linewidth]{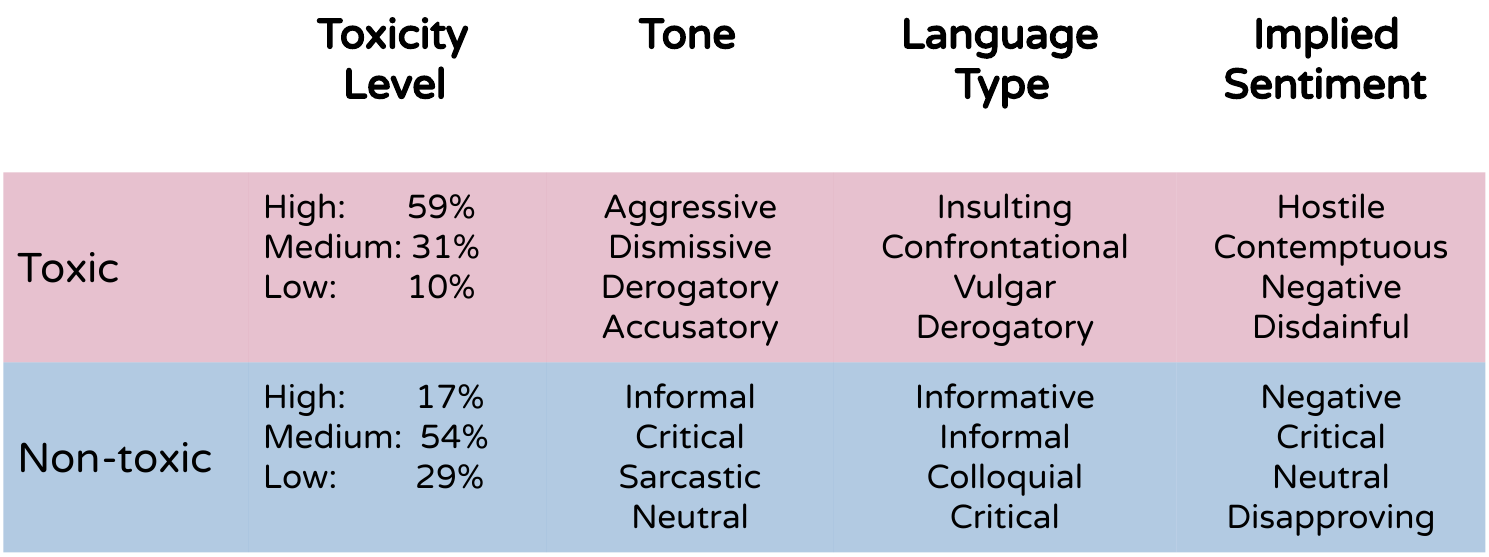}
        \vspace{0.5cm}
        \caption{Descriptive Features}
    \end{subfigure}
    \caption{Extracted with GPT-4 toxicity levels and top descriptive features per toxic and non-toxic parts in the multilingual parallel text detoxification data.}
    \label{fig:descriptive_features_toxicity_level}
\end{figure*}

\subsubsection{Annotation Process}

\paragraph{Annotation Setup} For data annotation and verification, we employed a specifically designed three-task pipeline:
\textit{Task 1: Determine if the sentences are toxic.} Annotators were required to choose one of three options: the given sentence is \textit{neutral, toxic but can be rewritten, or toxic and cannot be rewritten}. The last option was included based on the observation that some toxic texts are impossible to rewrite in a non-toxic manner.
\textit{Task 2: Rewrite sentences in a non-toxic style.} Annotators were instructed to create detoxified versions of the toxic sentences identified in Task 1 preserving the main content of the original sentences and rewriting the toxic words.
\textit{Task 3: Cross-check verification.} The detoxified sentences were assigned to different annotators for verification to ensure the quality.




\paragraph{Annotators} We hired three native Chinese annotators from mainland China—two 22-year-old women with Bachelor's degrees in engineering and one 32-year-old man with a Master's in computer science---ensuring strong familiarity with both the language and the detoxification task.



\subsection{Final Dataset}

The full picture of newly collected and available for now parallel detoxification data in 9 languages is presented in Table~\ref{tab:final_paradetox_dataset}. In the final stage, experts and native speakers thoroughly reviewed the entire dataset to ensure it met the task's specific requirements and criteria. 
Using both existing~\cite{logacheva-etal-2022-paradetox,dementieva2024multiparadetox} and newly collected data, we selected 1\,000 samples per language which were then split into 400 training and 600 test instances.\footnote{\href{https://huggingface.co/datasets/textdetox/multilingual_paradetox}{https://huggingface.co/datasets/textdetox/\\multilingual\_paradetox}} These datasets and their respective divisions were subsequently utilized for further described analysis and experiments.

\section{Explaining ParaDetox with LLM}

Although Large Language Models (LLMs) still have room for improvement in text classification tasks, specifically, for hate and toxic speech~\cite{roy-etal-2023-probing}, they have shown significant success in generating explanations~\cite{DBLP:journals/corr/abs-2402-01761}. Given the resource-intensive nature of manually annotating descriptive aspects for each sample across multiple languages, we utilized GPT-4 to assist in generating explanations. We ensured the quality of these explanations by validating them with native speakers, while also conducting an in-depth analysis of parallel text detoxification data.

\subsection{Approach}

For all our experiments, we employ GPT-4~\cite{openai2023chatgpt} (May, 2024) leveraging the Chain-of-Thought reasoning method~\cite{qiao-etal-2023-reasoning} and the CO-STAR framework~\cite{DBLP:journals/corr/abs-2112-00819} specifically designed for reasoning about toxicity and stereotypical biases in data to enhance the detoxification prompt design. All 1\,000 pairs per nine languages were used for this analysis. The full texts of all prompts are available in Appendix~\ref{sec:app_prompts}.

We compare toxic and detoxified parts to validate the detoxification process and identify cross-lingual similarities and differences in toxicity. For both parts, we extract descriptive features---toxicity level, tone, language type, implied sentiment, and negative connotation---using the following prompt (Appendix~\ref{sec:app_descriptive_prompt}, output example in Table~\ref{tab:parallel_examples_english}):\\
\textsf{\small \textbf{Sentence}: \{sentence\}; \\
\textbf{Toxicity Level}: Specify here (Low/Medium/High); \\
\textbf{Tone}: the overall tone of the sentence--choose from keywords; \\
\textbf{Language}: Language style--choose from keywords; \\
\textbf{Implied Sentiment}: the overall sentiment--choose from keywords; \\
\textbf{Context}: Brief description of how context contributes to toxicity; \\
\textbf{Negative Connotations}: List specific negative words/phrases here.
}

\begin{figure}[ht!]
    \centering
    \includegraphics[width=1\linewidth]{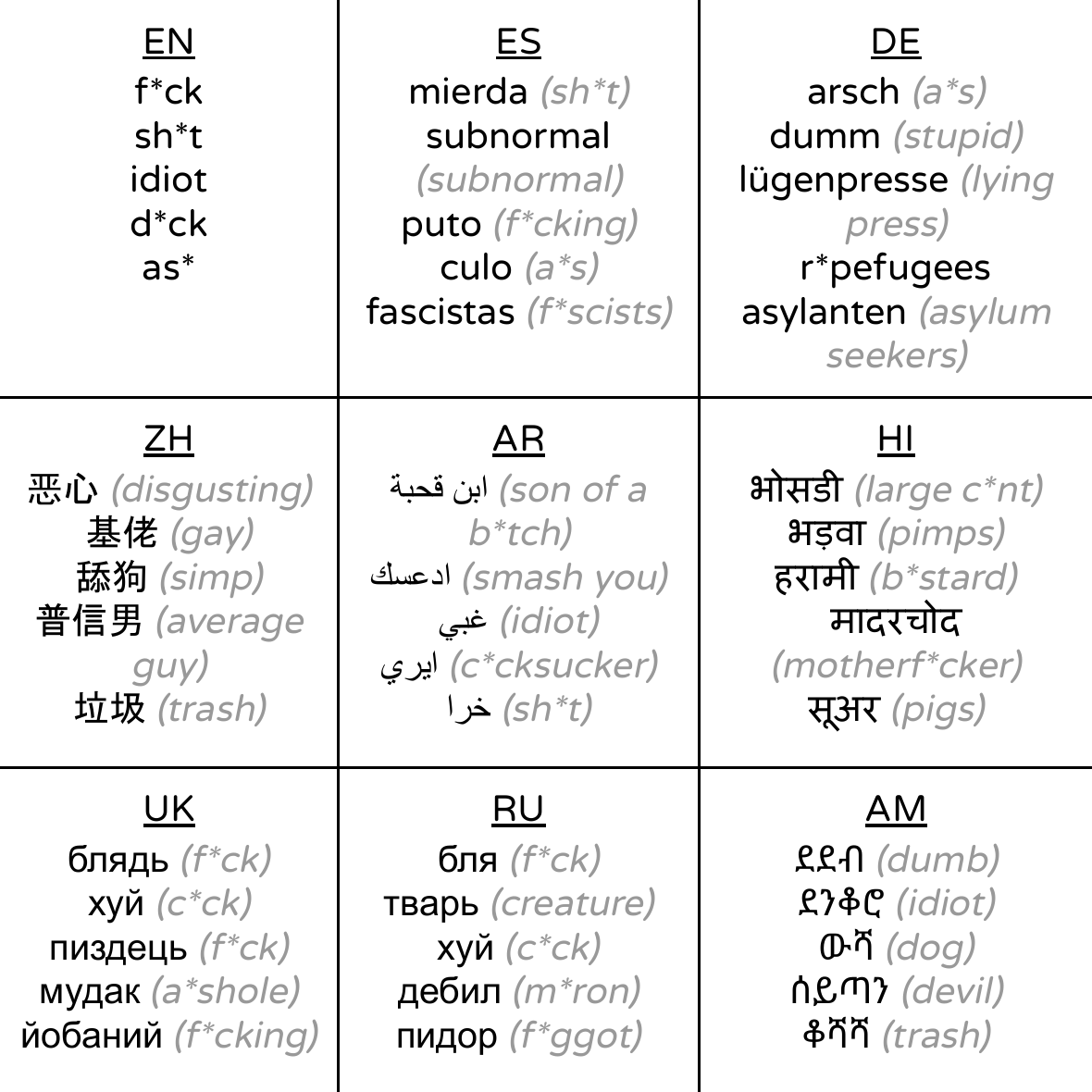}
    \caption{Top-5 extracted keywords from toxic parts.}
    \label{fig:toxic_keywords}
\end{figure}


We first prompted the model for open-ended descriptions for each feature, then selected the top 30 keywords from the explanations to refine the prompt, minimizing hallucinations. The core prompt was in English, with the target sentence in the respective language.
Experts and native speakers reviewed all 1\,000 samples per language for each feature and toxic keyword. All experts observed GPT-4’s tendency to overreact to certain keywords, yet its toxicity rankings were accurate. For descriptive features and toxic keywords across all languages, experts agreed with GPT-4’s answers in 98\% of cases.

\subsection{Toxicity Descriptive Features Analysis}

The overall view on top descriptive features for all languages as well as toxicity level per language are provided in Figure~\ref{fig:descriptive_features_toxicity_level}. The full list of top descriptive feature per language are provided in Appendix~\ref{sec:app_descriptive_features}.

Across all languages, we observe a reduction from high toxicity to medium or low levels, confirming that the paraphrases have been effectively detoxified. The original texts are predominantly \textit{aggressive}, \textit{derogatory}, \textit{vulgar}, and \textit{insulting}, often conveying \textit{hostile}, \textit{negative}, and \textit{disdainful} sentiments. In contrast, the neutral paraphrases tend to shift towards \textit{informal}, \textit{colloquial}, or even \textit{neutral} language, though they may still retain some \textit{negative} or \textit{critical} undertones.

\subsection{Toxic Keywords Analysis}

We extracted the most frequent toxic collocations from the toxic texts, as shown in Figure~\ref{fig:toxic_keywords}.

We found both similarities and differences in the typical rude and obscene language across languages. While some toxic words---like, \textit{f*ck}, \textit{idiot}, \textit{as*}---are present almost in all target languages, we can also see cultural specifics. In Ukrainian, Russian, and Chinese, derogatory comparisons involving homosexual individuals are considered insults, while in Hindi and Amharic, referring to someone using animal names is more prevalent. In Germany, while the issue of temporarily displaced individuals sparks significant societal debate, rudeness often manifests through wordplay targeting these individuals. As a result, while common obscene language appears across all languages, the expressions of toxicity are culturally dependent thus requires culture-aware toxicity mitigation solutions.

\subsection{Text Detoxification Analysis}

Then, we analyzed the way how detoxification was performed (see Table~\ref{tab:edit_types}). We sought lemmas that reflect various editorial actions---\textit{delete}, \textit{remove}, \textit{rephrase}, \textit{replace}, \textit{insert}, \textit{add}---using the following prompt template: \textsf{\small Answer shortly, how this text: \{toxic text\} was rephrased into this: \{detoxified text\}}. Additionally, we computed the Levenshtein distance between toxic and non-toxic parts (Appendix~\ref{sec:app_lengths}).

Across all languages, adding new content is rare. Detoxification mainly involves removing or rephrasing toxic elements. In German, Arabic, Hindi, Ukrainian, and Amharic, removal and rephrasing occur equally, while Spanish favors removal and Chinese/Russian rely more on rephrasing. Consequently, localized edits with fluent substitutions generally suffice for effective detoxification.

\begin{table}[hb!]
 \centering
 \footnotesize
  \setlength\tabcolsep{5pt}
 \begin{tabular}{l|lll|l|lll}
 \toprule
     Lang.  & Del    & Rep    & Ins  & Lang. & Del    & Rep    & Ins   \\ 
     \midrule
     EN & $27$\%     & $60$\%     & $13$\% & HI & $47$\%     & $44$\%     & $9$\%  \\
     ES & $60$\%     & $26$\%     & $14$\% & UK & $37$\%     & $59$\%     & $4$\%  \\
     DE & $44$\%     & $48$\%     & $8$\%  & RU & $23$\%     & $71$\%     & $6$\%  \\
     ZH & $14$\%     & $84$\%     & $2$\%  & AM & $45$\%     & $44$\%     & $11$\%  \\
     AR & $35$\%     & $55$\%     & $10$\%   \\
 \bottomrule
 \end{tabular}
 \caption{Percentage of toxic phrases \textbf{\underline{Del}}eted, \underline{\smash{\bf Rep}}hrased, or new non-toxic parts \textbf{\underline{Ins}}erted in order to achieve detoxification.}
 \label{tab:edit_types}
\end{table}

\subsection{Chain-of-Thoughts Text Detoxification}

\begin{figure}[ht!]
    \centering
    \includegraphics[width=0.9\linewidth]{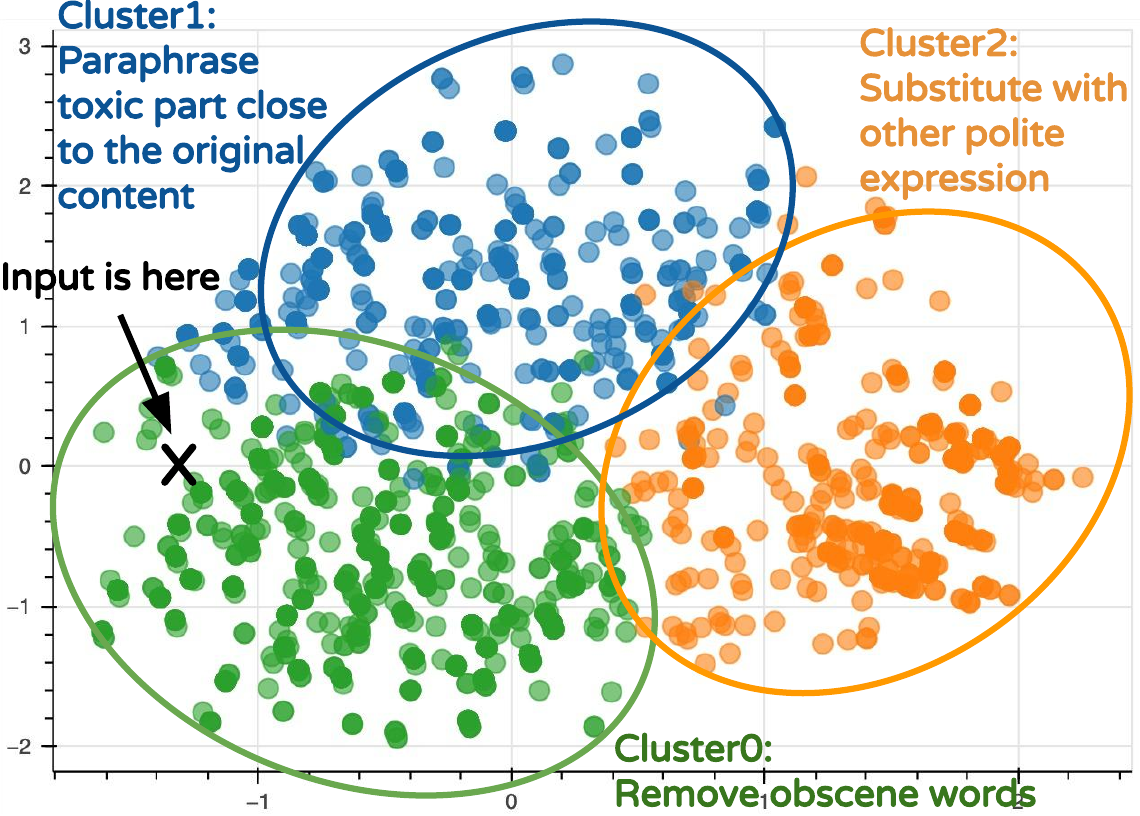}
    \caption{Text detoxification with CoT: analyze the input, identify its cluster, and provide the detoxification explanation and cluster example in the prompt.}
    \label{fig:detoxification_cot}
\end{figure}

Finally, we developed a new chain-of-thought reasoning approach to improve text detoxification with LLMs by guiding detoxification with explanations and close examples.

Our descriptive analysis suggests that the most effective detoxification approach varies according to descriptive features, toxicity expression and the target language itself. Depending on these factors, the detoxification strategy should be chosen accordingly. While it is challenging to come up with the clear human-readable instruction, the detoxification can be explained via examples. Thus, based on the extracted descriptive features, we performed K-means clustering per language on their one-hot encodings. The experiments with hyperparameters indicated an optimal division into $3$ clusters with the following top descriptive features and approximate explanations (Figure~\ref{fig:detoxification_cot}, i.e. for English):
\begin{itemize}[leftmargin=10pt]
\itemsep-0.2em  
    \item \textbf{Cluster $0$}: \textit{Offensive}, \textit{hostile}, and characterized by \textit{vulgar} language. Texts can be detoxified mainly by \textit{removing} profanities.
    \item \textbf{Cluster $1$}: \textit{Condescending}, \textit{derogatory}, \textit{dismissive}, and potentially \textit{biased} by gender or race. Here, texts requires more significant \textit{rephrasing} to remove condescending or biased language.
    \item \textbf{Cluster $2$}: \textit{Informal}, \textit{casual}, and \textit{playful}. Texts can be slightly adjusted by \textit{inserting} neutral or polite expressions after removing the toxic parts.
\end{itemize}
\vspace{-2mm}


Upon receiving new input, the LLM first estimates the descriptive features of a new text and the corresponding clustering is performed. LLM is then prompted to detoxify this sentence now using information about the cluster and a representative example of how to detoxify this type of cluster. The full prompt example can be found in Appendix~\ref{sec:app_cot_prompt} and clusters details in Appendix~\ref{sec:app_clusters}.

\section{Automatic Evaluation Setup}
We adopt the evaluation pipeline from \citet{logacheva-etal-2022-paradetox} to our multilingual setup. Direct links to the datasets/models instances are in Appendix~\ref{sec:app_auto_metrics}.

\paragraph{Style Transfer Accuracy (STA)}
We subsampled 5\,000 samples---2\,500 toxic and 2\,500 neutral---from toxicity classification corpora for each language (see in Table~\ref{tab:final_paradetox_dataset}) that were not used for ParaDetox data collection. We  fine-tuned XLM-R-\texttt{large}~\cite{DBLP:conf/acl/ConneauKGCWGGOZ20} instance for the binary toxicity classification task.

\paragraph{Content Similarity (SIM)} is the cosine similarity between LaBSE embeddings~\cite{DBLP:conf/acl/FengYCA022} of the source texts and the generated texts. 

\paragraph{Fluency (ChrF1)} is used to estimate the proximity of the detoxified texts to human references. we use an implementation of ChrF1 score from \texttt{sacrebleu} library~\cite{DBLP:conf/wmt/Post18}. 

\paragraph{Joint score (J)} is the aggregation of the three above metrics:
\begin{center}
    $\textbf{J} = \frac{1}{n}\sum\limits_{i=1}^{n}\textbf{STA}(y_i) \cdot \textbf{SIM}(x_i,y_i) \cdot \textbf{ChrF1}(x_i, y_i)$,
\end{center}
where \textbf{STA}($y_i$), \textbf{SIM}($x_i,y_i$), \textbf{ChrF1}($x_i,y_i$) $\in [0, 1]$ for each text detoxification output $y_i$.

\section{Baselines}

\begingroup
\renewcommand{\arraystretch}{1.15}
\begin{table*}[ht!]
    \centering
    \footnotesize
    \begin{tabular}{l|c| c c c c c c c c c}
    \toprule
         & \textbf{Average} & \textbf{EN} & \textbf{ES} & \textbf{DE} & \textbf{ZH} & \textbf{AR} & \textbf{HI} & \textbf{UK} & \textbf{RU} & \textbf{AM} \\  
        \midrule         
        \rowcolor{Lightgreen} Human References & $0.608$ & $0.711$ & $0.709$ & $0.733$ & $0.201$ & $0.695$ & $0.298$ & $0.790$ & $0.732$ &	$0.601$ \\
        \midrule
        & \multicolumn{10}{c}{\textbf{\textit{Unsupervised Approaches}}} \\
        \midrule
        Duplicate & $0.126$ & $0.061$ & $0.090$ & $0.287$ & $0.069$ & $0.294$ & $0.035$ & $0.032$ & $0.048$ & $0.217$ \\
        Delete & $\boldsymbol{0.302}$ & $0.447$ & $0.319$ & $\boldsymbol{0.362}$ & \underline{$\boldsymbol{0.175}$} & \underline{$\boldsymbol{0.456}$} & $\boldsymbol{0.105}$ & $\boldsymbol{0.328}$ & $\boldsymbol{0.255}$ & \underline{$\boldsymbol{0.270}$} \\ 
        Backtranslation & $0.205$ & \underline{$\boldsymbol{0.506}$} & $0.275$ & $0.233$ & $0.027$ & $0.206$ & $0.104$ & $0.201$ & $0.223$ & $0.075$ \\ 
        condBERT &  $0.213$ & $0.278$ & $\boldsymbol{0.347}$ & $0.310$ & $0.067$ & $0.337$ & $0.033$ & $0.316$ & $0.224$ & $0.003$\\
        \midrule
        & \multicolumn{10}{c}{\textbf{\textit{Supervised Approaches}}} \\
        \midrule
        mBART-Translated & $\boldsymbol{0.291}$ & $\boldsymbol{0.443}$ & $\boldsymbol{0.315}$ & $0.392$ & $\boldsymbol{0.083}$ & $0.365$ & $0.142$ & $0.343$ & $\boldsymbol{0.359}$ & $0.178$ \\
        mBART-mParaDetox & $0.282$ & $0.339$ & $0.289$ & \underline{$\boldsymbol{0.409}$} & $0.068$ & $\boldsymbol{0.397}$ & $\boldsymbol{0.171}$ & $\boldsymbol{0.345}$ & $0.321$ & $\boldsymbol{0.204}$ \\

        \midrule
        & \multicolumn{10}{c}{\textbf{\textit{LLM-based Approaches}}} \\
        \midrule
        GPT-4 few-shot & $0.324$ & $\boldsymbol{0.475}$ & $0.422$ & $0.396$ & $0.109$ & $0.270$ & $0.194$ & $0.460$ & $0.383$ & $\boldsymbol{0.205}$ \\  
        GPT-4 CoT (ours) & \underline{$\boldsymbol{0.331}$} & $0.326$ & \underline{$\boldsymbol{0.447}$} & $\boldsymbol{0.400}$ & $\boldsymbol{0.117}$ & $\boldsymbol{0.339}$ & $\boldsymbol{0.251}$ & \underline{$\boldsymbol{0.503}$} & \underline{$\boldsymbol{0.426}$} & $0.166$ \\
    \bottomrule
    \end{tabular}
    \caption{Results of the \emph{automatic} evaluation of the text detoxification approaches. The scores for each language are respective \textbf{J}oint scores. \textbf{Bold} denote the best results within the group, \underline{\textbf{underlined}}---the best for the language.}
    \label{tab:auto_results_final}
\end{table*}
\endgroup

For comparison, we considered several unsupervised and supervised text detoxification approaches together with a baseline prompt construction. Details of the hyperparameters and model choices for each method can be found in Appendix~\ref{sec:app_hyperparameters}.

\paragraph{Duplicate} Trivial baseline: the output sentence is a copy-paste of the input sentence. This baseline has $1.0$ (or $100\%$) SIM score by definition. 

\paragraph{Delete} Removal of offensive terms using a manually compiled list of vulgar words.
We collected and compiled together the lists of such toxic keywords for all target languages based on openly available sources (see Table~\ref{tab:multilingual_tox_keywords}).

\paragraph{Backtranslation} As for a more sophisticated unsupervised baseline, we performed translation of non-English texts into English with NLLB~\cite{DBLP:journals/corr/abs-2207-04672} to then perform detoxification with the fine-tuned on English ParaDetox BART~\cite{logacheva-etal-2022-paradetox}. The detoxification results were translated back to the target languages.

\paragraph{condBERT} We adapted one of the MLM-based unsupervised methods from~\citet{dale-etal-2021-text}. We used mBERT~\cite{devlin-etal-2019-bert} as a base model. The model runs MLM to generate list of substitutes selecting non-toxic ones.

\paragraph{Fine-tuned LM on Translated Data} We also tried to obtain synthetic parallel corpora by translating selected 400 English ParaDetox samples to our target languages. We utilized mBART for machine translation model~\cite{DBLP:journals/tacl/LiuGGLEGLZ20} for the translation step. We tuned the mBART for text generation~\cite{tang2020multilingual} on the obtained data.

\paragraph{Fine-tuning on the parallel data}  Finally, we fine-tuned the multilingual text-to-text generation model mBART-Large on the selected training multilingual data.

\paragraph{GPT-4 few-shot prompting} Before CoT, we applied a few-shot prompting of GPT-4 with the example prompt presented in Appendix~\ref{sec:app_few_shot_prompt}.

\section{Results}
We conducted a multilingual text detoxification across all languages on the test sets, with the results presented in Table~\ref{tab:auto_results_final} and detailed metrics per language in Appendix~\ref{sec:app_results}. Surprisingly, the Delete method outperformed other unsupervised approaches for three languages---Chinese, Arabic, and Amharic. This may be due to the nature of these languages (Table~\ref{tab:edit_types}), where detoxification relies heavily on paraphrasing. Since the proposed methods still struggled with appropriate paraphrasing, Delete, which removes toxic content without rephrasing, performed best. However, for other languages, where rephrasing is also key, LM-based solutions excelled, likely due to better representation of the languages in the pre-training data.

While for the majority of languages mBART fine-tuned on human-curated data outperformed the model fine-tuned on translated data, this results is not consistent. As described previously, some obscene terms are similar across languages and can be translated from English, offering sufficient information about toxicity for the target language. However, in the case of German, Hindi, Ukrainian, and Amharic cultural nuances play a significant role, leading the model trained on manually crafted data to perform better.

Finally, incorporating cluster information into the prompting process significantly boosted GPT-4 CoT's performance, surpassing the few-shot prompting approach for nearly all languages. This suggests that targeting toxicity with greater precision and information on relevant human-curated detoxifications reduces model hallucinations. As a result, this method achieved the highest scores across all approaches in the STA metric and standouts with the highest average J score (see example in Table~\ref{tab:output_examples_english}).

\section{Conclusion}

This work addressed the multilingual and explainability aspects of the text detoxification task. We introduced manually curated parallel detoxification datasets for new languages---German, Chinese, Arabic, Hindi, and Amharic---and the detailed data collection process. Next, we used LLMs as explainability tools on nine languages to analyze key descriptive features of toxic and non-toxic texts, identify top toxic collocations, and determine the primary actions required for detoxification per different toxicity expressions. Building on these insights, we developed a new Chain-of-Thoughts LLM prompting text detoxification method that incorporates detoxification cluster information about the input text. This approach reduced model's hallucinations, improved precision in edits, incorporated cultural specifics, and outperformed all baselines.

\section*{Limitations}
Firstly, while the work aims to extend data to new languages, there remains significant room for improvement in incorporating as many languages as possible. The selection of languages in this study was based on the native languages of the authors, but broader involvement of other language stakeholders could enhance the dataset.

Secondly, this work focuses solely on multilingual detoxification without exploring monolingual or cross-lingual tasks. Further research could be conducted to identify the most effective detoxification model for each language using the created data. Additionally, cross-lingual approaches could explore how detoxification knowledge transfers between languages, opening new avenues for research. Preliminary cross-lingual transfer experiments have been conducted for English and Russian~\cite{dementieva-etal-2023-exploring}, but the new dataset now includes more languages for further exploration.

For the CoT approach, we focused on human-readable cluster explanations in English; however, this approximation was not thoroughly explored for other languages. Our method currently relies on example-based explanations, and further research into human-readable cluster descriptions remains open for future work.

Lastly, the primary experiments in this study were conducted using GPT-4, a closed-source model from OpenAI. While GPT-4 continues to perform exceptionally well in various NLP benchmarks, demonstrating stable generation of coherent explanations, we recognize the importance of supporting open-source initiatives. Therefore, we acknowledge the necessity of ablation study with opensource LLMs.

\section*{Ethics Statement}

We explore the task of text detoxification with no intent to violate the freedom of speech, but rather to help mitigate digital violence, create safer online environments for children, and promote the development of secure AI models. The ideal implementation of detoxification models on communication platforms would be as suggestions, rather than forced corrections. A user-friendly interface for these suggestions should be considered by stakeholders.

Additionally, detoxifying LLMs, not just human content, is a relevant topic. Already several approaches were explored~\cite{DBLP:conf/emnlp/LeongCWWL23,DBLP:journals/corr/abs-2406-07973} utilizing English ParaDetox data as instruction dataset to mitigate toxicity in the model. However, these efforts have been limited to monolingual contexts due to data constraints. Further research into detoxifying LLMs in other languages, as well as the potential for cross-lingual knowledge transfer, represents a promising area for future study.

Finally, the authors of this work utilized ChatGPT to check the grammar and correct the appropriateness of the used language.

\section*{Acknowledgements}
This work was only possible to the massive support from various institutions. Firstly, DD, NR, SMY, DM, AM and AP would like to thank SPARC-II (Scheme for Promotion of Academic and Research Collaboration, Phase II) project for funding international travel and subsistence to carry out this work. Then, pilot experiments and additional annotation was supported by Toloka.ai research grant. The further contribution of DD of this work was supported by the Friedrich Schiedel Fellowship hosted by the TUM School of Social Sciences and Technology and the TUM Think Tank. We sincerely acknowledge the financial support provided by the fellowship. Additionally, we would like to extend our gratitude to the TUM Data Analytics\&Statistics chair, under the leadership of Alexander Fraser.


\bibliography{anthology,custom}

\appendix

\onecolumn

\section{Prompts for Explanations and Chain-of-Thoughts Detoxification with LLMs}
\label{sec:app_prompts}

Here, we provide exact prompts used for explaining multilingual parallel detoxification data and text detoxification prompting.



\subsection{Prompt for Descriptive Features Extraction}
\label{sec:app_descriptive_prompt}

\begin{tcolorbox}
\textbf{Please analyze the provided sentence using the structure below to identify elements of toxicity and suggest improvements},
when I tell you, use words from the keywords list (can be more than one word!): 

\textbf{keywords} = [Neutral, Informative, Casual, Assertive, Dismissive, Condescending, \\
Friendly, Commanding, Instructive  Derogatory, Confrontational, Insulting, \\
Vulgar, Formal, Informal, Offensive, Technical, Playful, Positive, \\
Frustration, Analytical, Professional, Hostile, Hatred, Helpful, \\
Angry, Friendly, Arrogant]
\\
\textbf{Analysis Structure} (do not use '' and [] and "" in your answer and do not suggest improvement!):

\{ \\
\hspace*{0.5cm}\textit{Sentence}: \{sentence\}, \\
\hspace*{0.5cm}\textit{Toxicity Level}: Specify here (Low/Medium/High), \\
\hspace*{0.5cm}\textit{Tone}: the overall tone of the sentence- choose from keywords, \\
\hspace*{0.5cm}\textit{Language}: Language style---choose from keywords, \\
\hspace*{0.5cm}\textit{Implied Sentiment}: the overall sentiment- choose from keywords, \\
\hspace*{0.5cm}\textit{Context}: Brief description of how context contributes to toxicity, \\
\hspace*{0.5cm}\textit{Negative Connotations}: List specific negative words/phrases here, \\
\hspace*{0.5cm}\textit{Intent}: Describe the perceived intent behind the sentence. \\
\}
\end{tcolorbox}

\subsection{Few-Shot Prompting for Text Detoxification}
\label{sec:app_few_shot_prompt}

\begin{tcolorbox}
\textbf{Please detoxify the provided sentence} using the structure below without changing the real meaning of the sentence.\\
\textbf{Analysis Structure} (do not use '' and [] and "" in your answer and do not suggest improvement!):
\{ \\
\hspace*{0.5cm}\textit{Sentence}: \{sentence\}, \\
\hspace*{0.5cm}\textit{Fixed sentence}: $<$the non-toxic sentence without changing the meaning$>$ \\
\},

\textbf{Example 1}:
\{ \\
\hspace*{0.5cm}\textit{Sentence}: dude should have been taken to api , he would be right at home with all the other knuckleheads there, \\
\hspace*{0.5cm}\textit{Fixed sentence}: It would have been good if he went to api. He would fit in. \\
\} \\
\textbf{Example 2}:
\{ \\
\hspace*{0.5cm}\textit{Sentence}: damn those young mothers driving their children to daycare through the snow drifts., \\
\hspace*{0.5cm}\textit{Fixed sentence}: those young mothers driving their children to daycare through the snow drifts. \\
\}
\end{tcolorbox}






\subsection{Chain-of-Thoughts Prompting with Cluster Knowledge Incorporation}
\label{sec:app_cot_prompt}

\begin{tcolorbox}
\textbf{Please detoxify the provided sentence} using the structure below without changing the real meaning of the sentence. \\
The sentences are \textbf{clustered into 3 groups} while each group has its own characterizes. \\
\hspace*{0.5cm}\textbf{Cluster 0} is more Offensive, Hostile and Vulgar; \\
\hspace*{0.5cm}\textbf{Cluster 1} is more Condescending, Derogatory and Hostile; \\
\hspace*{0.5cm}\textbf{Cluster 2} is more Informal, Casual, Dismissive. \\

For each sentence and cluster that I give, \textbf{make the sentence non-toxic by making it Neutral/Informal/Casual without changing the meaning.} \\
\textbf{Analysis Structure} (do not use '' and [] and "" in your answer and do not suggest improvement!): \\
\{ \\
\hspace*{0.5cm}\textit{Sentence}: \{sentence\}, \\
\hspace*{0.5cm}\textit{Toxicity level}: \{Specify here\}, \\
\hspace*{0.5cm}\textit{Cluster}: \{cluster\}, \\
\hspace*{0.5cm}\textit{Fixed sentence}: $<$the non-toxic sentence after making it Neutral/Informal/Casual without changing the meaning$>$; \\
\}, \\

\textbf{Example}: \\
\{ \\
\hspace*{0.5cm}\textit{Sentence}: dude should have been taken to api , he would be right at home with all the other knuckleheads there, \\
\hspace*{0.5cm}\textit{Toxicity Level}: Medium, \\
\hspace*{0.5cm}\textit{Cluster}: 0, \\
\hspace*{0.5cm}\textit{Fixed sentence}: It would have been good if he went to api. He would fit in. \\
\} \\
\end{tcolorbox}


\section{Automated Evaluation Metrics Models}
\label{sec:app_auto_metrics}

The direct links to the datasets and models instances used for the evaluation setup:
\begin{itemize}
    \item The compiled toxicity binary classification dataset to fine-tune an STA classifier;~\footnote{\href{https://huggingface.co/datasets/textdetox/multilingual_toxicity_dataset}{https://huggingface.co/datasets/textdetox/multilingual\_toxicity\_dataset}}

    \item The fine-tuned XLM-RoBERTa for STA estimation;~\footnote{\href{https://huggingface.co/textdetox/xlmr-large-toxicity-classifier}{https://huggingface.co/textdetox/xlmr-large-toxicity-classifier}}
    
    \item LaBSE multilingual encoder for SIM metric.\footnote{\href{https://huggingface.co/sentence-transformers/LaBSE}{https://huggingface.co/sentence-transformers/LaBSE}} 

\end{itemize}

\section{Hyperparameters Configurations for Considered Text Detoxification Approaches}
\label{sec:app_hyperparameters}

Here, we provide the final hyperparameters and other details for the main considered text detoxification baselines, fine-tuned multilingual text generation models, and GPT-4.

\subsection{Delete}

The resources used for the multilingual toxicity lexicon compilation are listed in Table~\ref{tab:multilingual_tox_keywords}. The full list is available online for public usage and reproducibility.\footnote{\href{https://huggingface.co/datasets/textdetox/multilingual_toxic_lexicon}{huggingface.co/datasets/textdetox/multilingual\_toxic\_lexicon}}

\begin{table}[ht!]
    \centering
    \footnotesize
    \begin{tabular}{l|c|r}
    \toprule
       \textbf{Language} & \textbf{Original Source} & \textbf{\# of Keywords} \\
    \midrule
       English & \cite{logacheva-etal-2022-paradetox, entoxkeywords,DBLP:journals/corr/abs-2207-04672} & 3\,390\\
       Spanish & \cite{DBLP:journals/corr/abs-2207-04672} & 1\,200 \\
       German & \cite{multitoxkeywords,DBLP:journals/corr/abs-2207-04672} & 247 \\
       Chinese & \cite{DBLP:journals/osnm/JiangYLZ22,lu-etal-2023-facilitating,DBLP:journals/corr/abs-2207-04672} & 3\,840 \\
       Arabic & Ours+\cite{DBLP:journals/corr/abs-2207-04672} & 430 \\
       Hindi & \cite{DBLP:journals/corr/abs-2207-04672} & 133 \\
       Ukrainian & \cite{uktoxkeywords,DBLP:journals/corr/abs-2207-04672} & 7\,360\\
       Russian & \cite{Dementieva2022RUSSE2022FO,DBLP:journals/corr/abs-2207-04672} & 141\,000 \\
       Amharic & Ours+\cite{DBLP:journals/corr/abs-2207-04672} & 245 \\
    \bottomrule
    \end{tabular}
    \caption{The list of the original sources and the corresponding amount of obscene keywords used to compile multilingual toxic lexicon list for our Delete baseline.}
    \label{tab:multilingual_tox_keywords}
\end{table}

\subsection{Backtranslation}

For the translation step, we used the NLLB instance.\footnote{\href{https://huggingface.co/facebook/nllb-200-distilled-600M}{https://huggingface.co/facebook/nllb-200-distilled-600M}} For English sentences detoxification, we utilized previously released BART-detox English instance.~\footnote{\href{https://huggingface.co/s-nlp/bart-base-detox}{https://huggingface.co/s-nlp/bart-base-detox}}

\subsection{condBERT}

We re-used of the condBERT pipeline introduced in~\cite{dale-etal-2021-text}~\footnote{\href{https://github.com/s-nlp/detox/tree/main/emnlp2021/style_transfer/condBERT}{https://github.com/s-nlp/detox/tree/main/emnlp2021/style\_transfer/condBERT}} with mBERT-base\footnote{\href{https://huggingface.co/google-bert/bert-base-multilingual-cased}{https://huggingface.co/google-bert/bert-base-multilingual-cased}} model and the hyperparameters for the masked language modelling task via \texttt{MaskedTokenPredictorBert} class with parameters \texttt{max\_len}$=250$ and \texttt{contrast\_penalty}$=0.0$.

\subsection{mBART}

Previous experiments in \citet{dementieva2024multiparadetox} showed quite poor performance of BloomZ-7b~\cite{DBLP:conf/acl/MuennighoffWSRB23} for the text detoxification. To choose the model for supervised fine-tuning for new multilingual text detoxification, we compared in this case two multilingual text generation models---mT0-large~\cite{DBLP:conf/acl/MuennighoffWSRB23}\footnote{\href{https://huggingface.co/bigscience/mt0-xxl-mt}{https://huggingface.co/bigscience/mt0-xxl-mt}} and mBART-large~\cite{tang2020multilingual}\footnote{\href{https://huggingface.co/facebook/mbart-large-50}{https://huggingface.co/facebook/mbart-large-50}}. The results comparison based on the overall J scores per language is presented in Table~\ref{tab:auto_results_text_generators}. In the end, for the final results, we chose mBART fine-tuned with the following setup: \texttt{num\_train\_epochs}$=10$, \texttt{warmusteps}$=10$, \texttt{learning\_rate}$=1e-05$, \texttt{batch\_size}$=32$. For the inference, we used the default parameters of \texttt{MBartForConditionalGeneration} class: \texttt{beams\_number}$=5$, \texttt{maximal\_tokens}$=200$.


\begingroup
\renewcommand{\arraystretch}{1.15}
\begin{table*}[h!]
    \centering
    \footnotesize
    \begin{tabular}{l|c| c c c c c c c c c}
    \toprule
         & \textbf{Average} & \textbf{EN} & \textbf{ES} & \textbf{DE} & \textbf{ZH} & \textbf{AR} & \textbf{HI} & \textbf{UK} & \textbf{RU} & \textbf{AM} \\  
        \midrule         
        \rowcolor{Lightgreen} Human References & $0.608$ & $0.711$ & $0.709$ & $0.733$ & $0.201$ & $0.695$ & $0.298$ & $0.790$ & $0.732$ &	$0.601$ \\
        \midrule
        & \multicolumn{10}{c}{\textbf{\textit{Supervised Approaches}}} \\
        \midrule
        mT0-Translated & $\boldsymbol{0.261}$ & $\boldsymbol{0.467}$ & $\boldsymbol{0.341}$ & $\boldsymbol{0.356}$ & $\boldsymbol{0.073}$ & $0.331$ & $0.106$ & $\boldsymbol{0.254}$ & $\boldsymbol{0.283}$ & $\boldsymbol{0.142}$ \\
        mT0-mParaDetox & $0.168$ & $0.397$ & $0.107$ & $0.244$ & $0.002$ & $\boldsymbol{0.356}$ & $\boldsymbol{0.150}$ & $0.040$ & $0.119$ & $0.097$ \\
        \midrule
        mBART-Translated & $\boldsymbol{0.291}$ & $\boldsymbol{0.443}$ & $\boldsymbol{0.315}$ & $0.392$ & $\boldsymbol{0.083}$ & $0.365$ & $0.142$ & $0.343$ & $\boldsymbol{0.359}$ & $0.178$ \\
        mBART-mParaDetox & $0.282$ & $0.339$ & $0.289$ & $\boldsymbol{0.409}$ & $0.068$ & $\boldsymbol{0.397}$ & $\boldsymbol{0.171}$ & $\boldsymbol{0.345}$ & $0.321$ & $\boldsymbol{0.204}$ \\
    \bottomrule
    \end{tabular}
    \caption{Results of the \emph{automatic} evaluation of the text detoxification approaches. The scores for each language are respective \textbf{J}oint scores. \textbf{Bold} denote the best results within the group.}
    \label{tab:auto_results_text_generators}
\end{table*}
\endgroup

\subsection{GPT-4 Prompting}

We employed GPT-4~\cite{openai2023chatgpt} for analysis and experiments during May, 2024. We used default hyperparameters for the inference step which included \texttt{temperature}$=1.0$, \texttt{top\_p}$=1.0$, \texttt{top\_k}$=0.0$, \texttt{frequency\_penalty}$=0.0$, \texttt{presence\_penalty}=$0.0$.

\section{Toxic and Detoxified Sentences Lengths Comparison}
\label{sec:app_lengths}
\begin{figure}[ht!]
    \centering
    \includegraphics[width=0.8\linewidth]{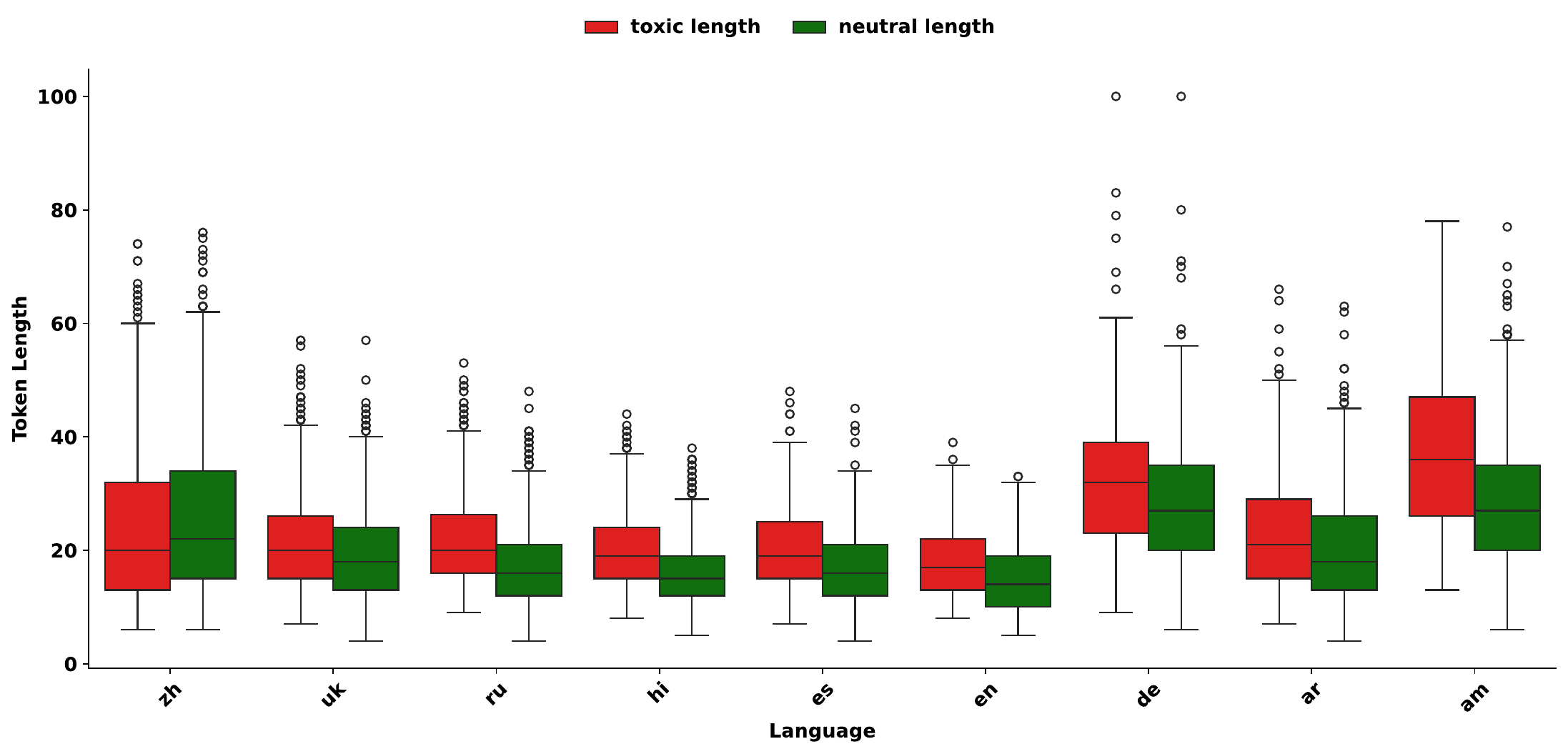}
    \caption{Comparison of toxic and non-toxic texts lengths distributions per each language.}
    \label{fig:lengths_comparison}
\end{figure}

\begin{figure}[ht!]
    \centering
    \includegraphics[width=0.8\linewidth]{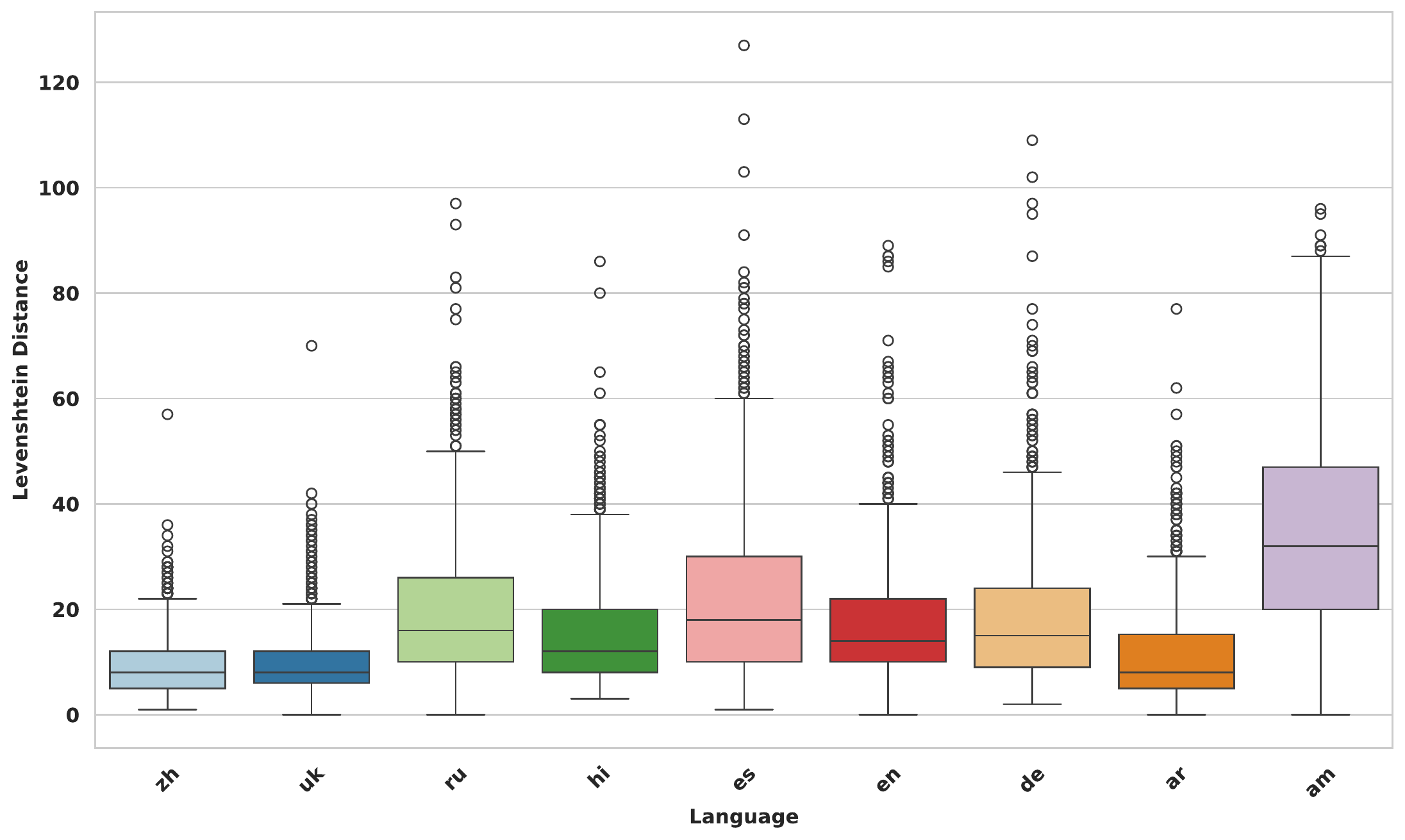}
    \caption{Levenshtein distances between toxic and non-toxic parts distribution.}
    \label{fig:levenshtein_distances}
\end{figure}

Additionally to the toxic keywords and edits types analysis, we also provide the lengths comparison of toxic and non-toxic parallel pairs (Figure~\ref{fig:lengths_comparison}) and the Levenshtein distances between them (Figure~\ref{fig:levenshtein_distances}). The lengths and distances calculation are based on the tokenization performed with \texttt{textdetox/xlmr-large-toxicity-classifier} used for STA calculation. Here, we again observe language-specific differences. For instance, in Chinese, detoxified versions are longer than their toxic counterparts, while in Amharic the length disparity is substantial. Even though toxic phrases are removed, the size of the replacement phrases can vary depending on both the language and the nature of the toxicity.

\newpage

\section{Top Descriptive Features}
\label{sec:app_descriptive_features}

\begingroup
\renewcommand{\arraystretch}{1.15}
\begin{table*}[ht!]
    \centering

    \definecolor{lightred}{rgb}{1, 0.9, 0.9}
    \definecolor{lightblue}{rgb}{0.8, 0.9, 1.0}
    
    \footnotesize
    \hspace*{-1cm}
    \begin{tabular}{>{\columncolor{lightred}}c|>{\columncolor{lightred}}c|>{\columncolor{lightred}}c|>{\columncolor{lightred}}c|c|>{\columncolor{lightblue}}c|>{\columncolor{lightblue}}c|>{\columncolor{lightblue}}c|>{\columncolor{lightblue}}c}
    \toprule
        \shortstack{\bf Toxicity \\ \bf Level} & \textbf{Tone} & \shortstack{\bf Language \\ \bf Type} & \shortstack{\bf Implied \\ \bf Sentiment} & \textbf{Language} & \shortstack{\bf Implied \\ \bf Sentiment} & \shortstack{\bf Language \\ \bf Type} & \shortstack{\bf Tone} & \shortstack{\bf Toxicity \\ \bf Level} \\
        \midrule
        
        \shortstack{High: 52\% \\ Medium: 38\% \\ Low: 10\% \\ \ } & \shortstack{Aggressive \\ Frustrated \\ Dismissive \\ Derogatory} & \shortstack{Vulgar \\ Insulting \\ Confrontat. \\ Informal} & \shortstack{Hostile \\ Negative \\ Angry \\ Critical} & \shortstack{\ \\ \textbf{EN} \\ \ } & \shortstack{Negative \\ Neutral \\ Frustrat. \\ Positive} & \shortstack{Informal \\ Informative \\ Direct \\ Critical} & \shortstack{Informal \\ Critical \\ Neutral \\ Accusatory} & \shortstack{High: 6\% \\ Medium: 51\% \\ Low: 43\% \\ \ } \\
        \midrule
        
        \shortstack{High: 35\% \\ Medium: 47\% \\ Low: 17\% \\ \ } & \shortstack{Aggressive \\ Frustrated \\ Dismissive \\ Insulting} & \shortstack{Vulgar \\ Insulting \\ Informal \\ casual} & \shortstack{Hostile \\ Negative \\ Contempt. \\ Angry} & \shortstack{\ \\ \textbf{ES} \\ \ } & \shortstack{Negative \\ Neutral \\ Frustrat. \\ Positive} & \shortstack{Informal \\ Informative \\ Colloquial \\ Neutral} & \shortstack{Informal \\ Neutral \\ Sarcastic \\ Critical} & \shortstack{High: 4\% \\ Medium: 43\% \\ Low: 53\% \\ \ } \\
        \midrule

        \shortstack{High: 70\% \\ Medium: 25\% \\ Low: 5\% \\ \ } & \shortstack{Aggressive \\ Dismissive \\ Derogatory \\ Accusatory} & \shortstack{Insulting \\ Derogatory \\ Confrontat. \\ Offensive} & \shortstack{Hostile \\ Negative \\ Angry \\ Disdainful} & \shortstack{\ \\ \textbf{DE} \\ \ } & \shortstack{Negative \\ Critical \\ Disapprov. \\ Disparaging} & \shortstack{Informal \\ Informative \\ Colloquial \\ Neutral} & \shortstack{Informal \\ Sarcastic \\ Accusatory \\ Critical} & \shortstack{High: 32\% \\ Medium: 57\% \\ Low: 11\% \\ \ } \\
        \midrule

        \shortstack{High: 45\% \\ Medium: 35\% \\ Low: 20\% \\ \ } & \shortstack{Dismissive \\ Derogatory \\ Aggressive \\ Neutral} & \shortstack{Insulting \\ Derogatory \\ Confrontat. \\ Casual} & \shortstack{Hostile \\ Contempt. \\ Negative \\ Disdainful} & \shortstack{\ \\ \textbf{ZH} \\ \ } & \shortstack{Negative \\ Critical \\ Disapprov. \\ Dismissive} & \shortstack{Informative \\ Informal \\ Critical \\ Derogatory} & \shortstack{Informal \\ Critical \\ Sarcastic \\ Neutral} & \shortstack{High: 45\% \\ Medium: 48\% \\ Low: 7\% \\ \ } \\
        \midrule

        \shortstack{High: 65\% \\ Medium: 25\% \\ Low: 10\% \\ \ } & \shortstack{Aggressive \\ Insulting \\ Dismissive \\ Accusatory} & \shortstack{Insulting \\ Confrontat. \\ Offensive \\ Derogatory} & \shortstack{Hostile \\ Contempt. \\ Negative \\ Disrespectful} & \shortstack{\ \\ \textbf{AR} \\ \ } & \shortstack{Negative \\ Critical \\ Neutral \\ Hostile} & \shortstack{Informative \\ Critical \\ Informal \\ Colloquial} & \shortstack{Critical \\ Informal \\ Accusatory \\ Sarcastic} & \shortstack{High: 20\% \\ Medium: 56\% \\ Low: 24\% \\ \ } \\
        \midrule

        \shortstack{High: 76\% \\ Medium: 18\% \\ Low: 6\% \\ \ } & \shortstack{Aggressive \\ Derogatory \\ Insulting \\ Accusatory} & \shortstack{Insulting \\ Offensive \\ Derogatory \\ Vulgar} & \shortstack{Hostile \\ Contempt. \\ Disrespectful \\ Negative } & \shortstack{\ \\ \textbf{HI} \\ \ } & \shortstack{Negative \\ Hostile \\ Informal \\ Aggressive} & \shortstack{Informal \\ Colloquial \\ Critical \\ Informative} & \shortstack{Accusatory \\ Critical \\ Informal \\ Aggressive} & \shortstack{High: 22\% \\ Medium: 63\% \\ Low: 15\% \\ \ } \\
        \midrule

        \shortstack{High: 61\% \\ Medium: 32\% \\ Low: 7\% \\ \ } & \shortstack{Aggressive \\ Frustrated \\ Dismissive \\ Casual} & \shortstack{Vulgar \\ Insulting \\ Confrontat. \\ Offensive} & \shortstack{Hostile \\ Negative \\ Angry \\ Contempt.} & \shortstack{\ \\ \textbf{UK} \\ \ } & \shortstack{Negative \\ Neutral \\ Frustration \\ Dismissive} & \shortstack{Colloquial \\ Informal \\ Informative \\ Conversat.} & \shortstack{Informal \\ Neutral \\ Casual \\ Sarcastic} & \shortstack{High: 5\% \\ Medium: 37\% \\ Low: 78\% \\ \ } \\
        \midrule

        \shortstack{High: 73\% \\ Medium: 22\% \\ Low: 5\% \\ \ } & \shortstack{Aggressive \\ Dismissive \\ Insulting \\ Derogatory} & \shortstack{Insulting \\ Confrontat. \\ Offensive \\ Vulgar} & \shortstack{Hostile \\ Contempt. \\ Negative \\ Disdainful} & \shortstack{\ \\ \textbf{RU} \\ \ } & \shortstack{Negative \\ Critical \\ Neutral \\ Disapprov.} & \shortstack{Informative \\ Colloquial \\ Informal \\ Critical} & \shortstack{Informal \\ Critical \\ Accusatory \\ Sarcastic} & \shortstack{High: 9\% \\ Medium: 64\% \\ Low: 27\% \\ \ } \\
        \midrule

        \shortstack{High: 55\% \\ Medium: 41\% \\ Low: 4\% \\ \ } & \shortstack{Aggressive \\ Accusatory \\ Derogatory \\ Critical} & \shortstack{Insulting \\ Confrontat. \\ Derogatory \\ Critical} & \shortstack{Hostile \\ Contempt. \\ Disapprov. \\ Negative} & \shortstack{\ \\ \textbf{AM} \\ \ } & \shortstack{Negative \\ Disapprov. \\ Critical \\ Neutral} & \shortstack{Critical \\ Informal \\ Accusatory \\ Confrontat.} & \shortstack{Critical \\ Accusatory \\ Informal \\ Confrontat.} & \shortstack{High: 14\% \\ Medium: 62\% \\ Low: 24\% \\ \ } \\
    \bottomrule
    \end{tabular}
    \caption{Main descriptive features per language for \textcolor{red}{toxic} (on the left) and \textcolor{blue}{detoxified} (on the right) parts.}
    \label{tab:discriptive_features}
\end{table*}
\endgroup


\begin{figure*}
        \centering
        \begin{subfigure}[b]{0.475\textwidth}
            \centering
            \includegraphics[width=\textwidth]{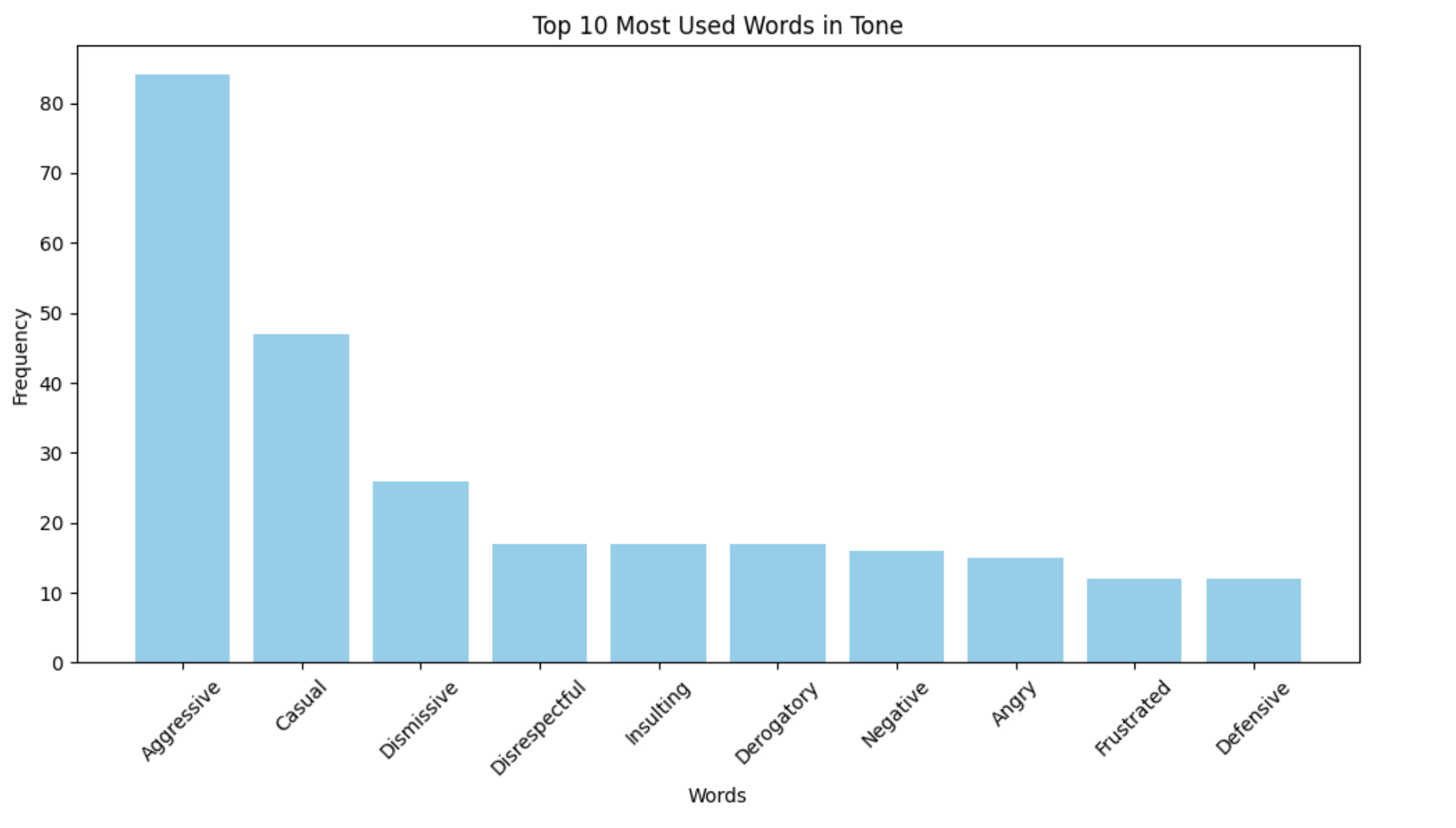}
            \caption[]%
           {{\small Tone}}    
            \label{tone}
        \end{subfigure}
        \hfill
        \begin{subfigure}[b]{0.475\textwidth}  
            \centering 
            \includegraphics[width=\textwidth]{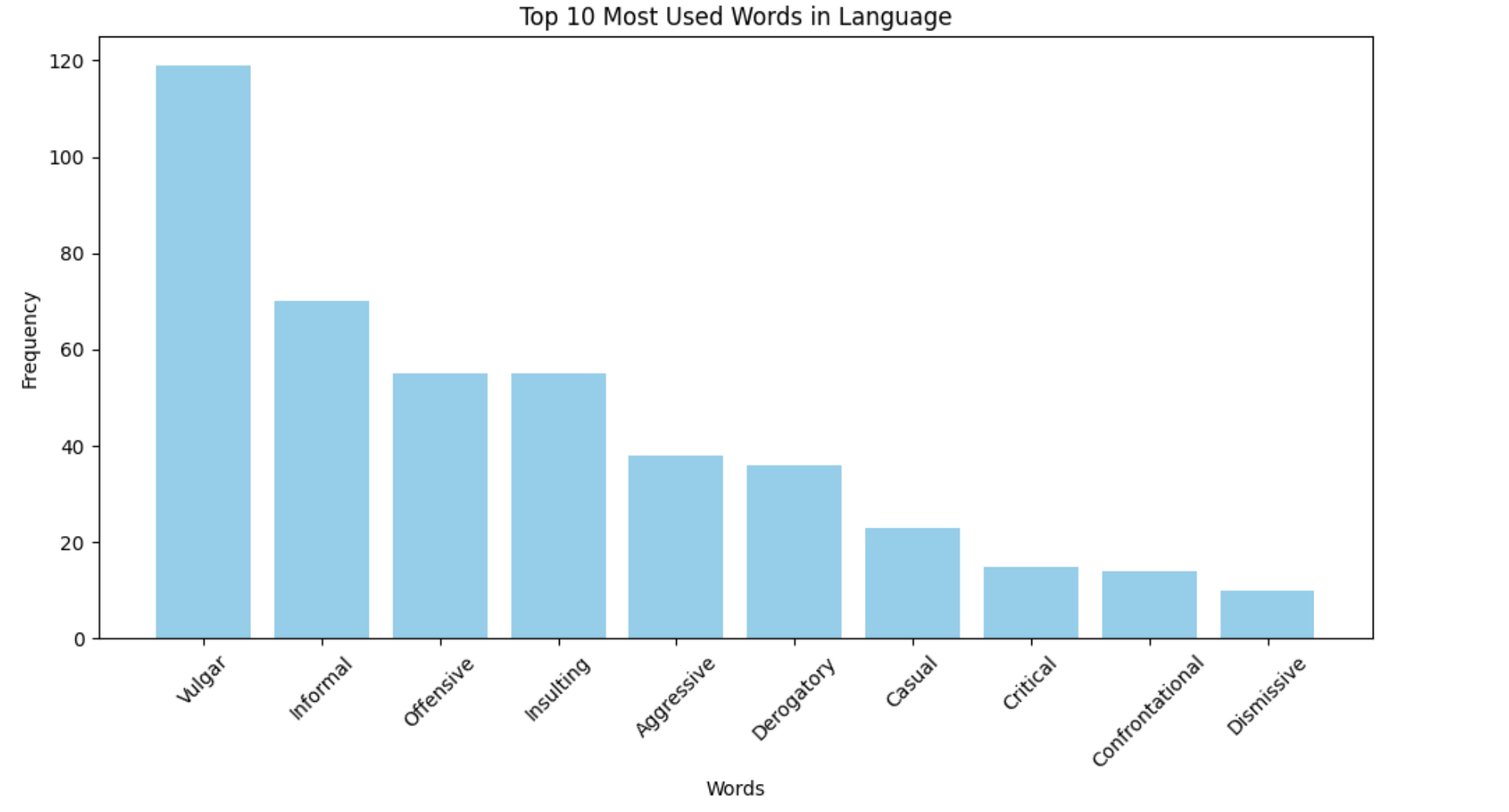}
            \caption[]%
           {{\small Language Type}}    
            \label{fig:cluster2}
        \end{subfigure}
        \vskip\baselineskip
        \begin{subfigure}[b]{0.475\textwidth}   
            \centering 
            \includegraphics[width=\textwidth]{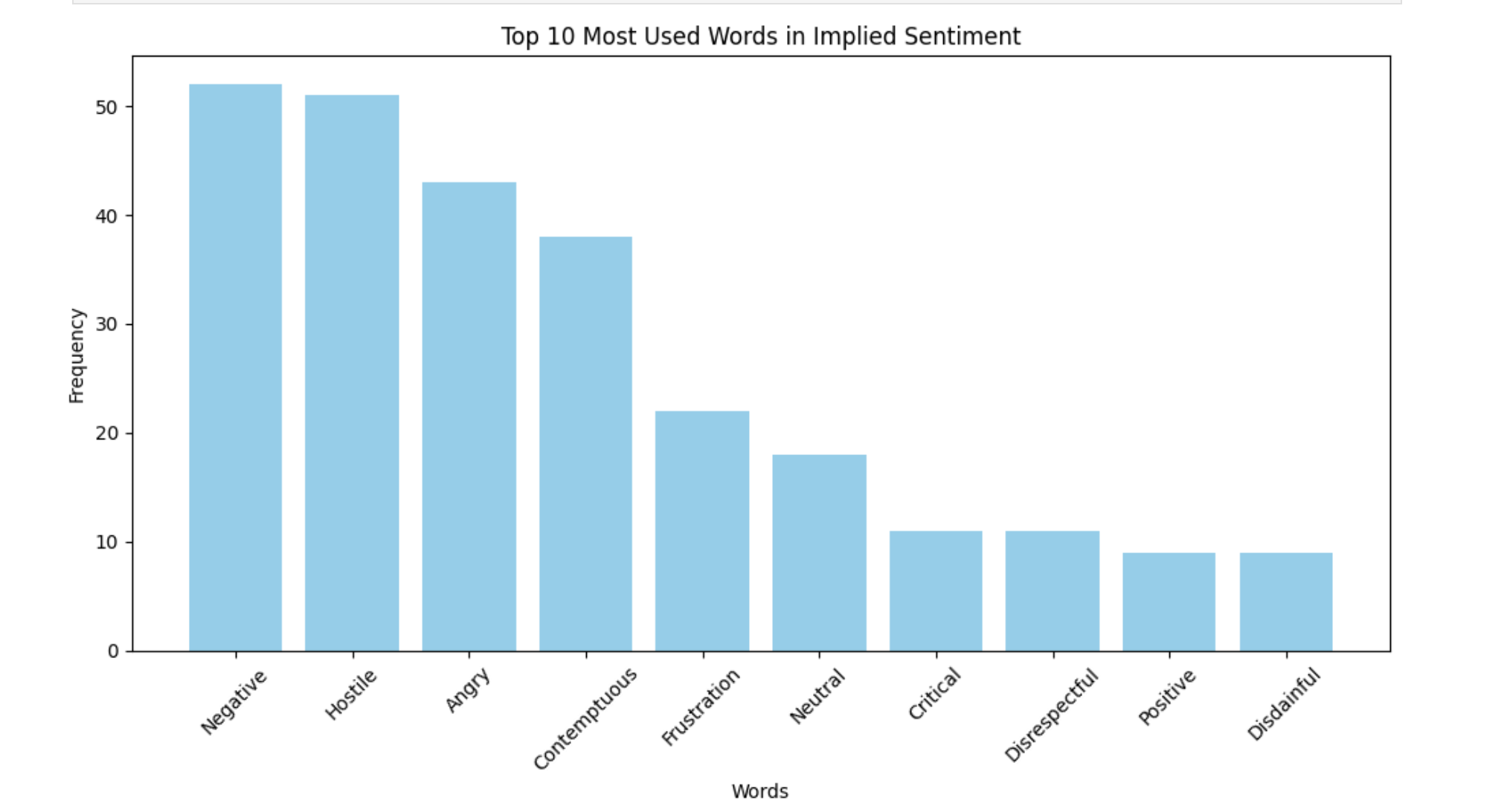}
            \caption[]%
           {{\small Sentiment}}    
            \label{fig:cluster3}
        \end{subfigure}
        \hfill
        \begin{subfigure}[b]{0.475\textwidth}   
            \centering 
            \includegraphics[width=\textwidth]{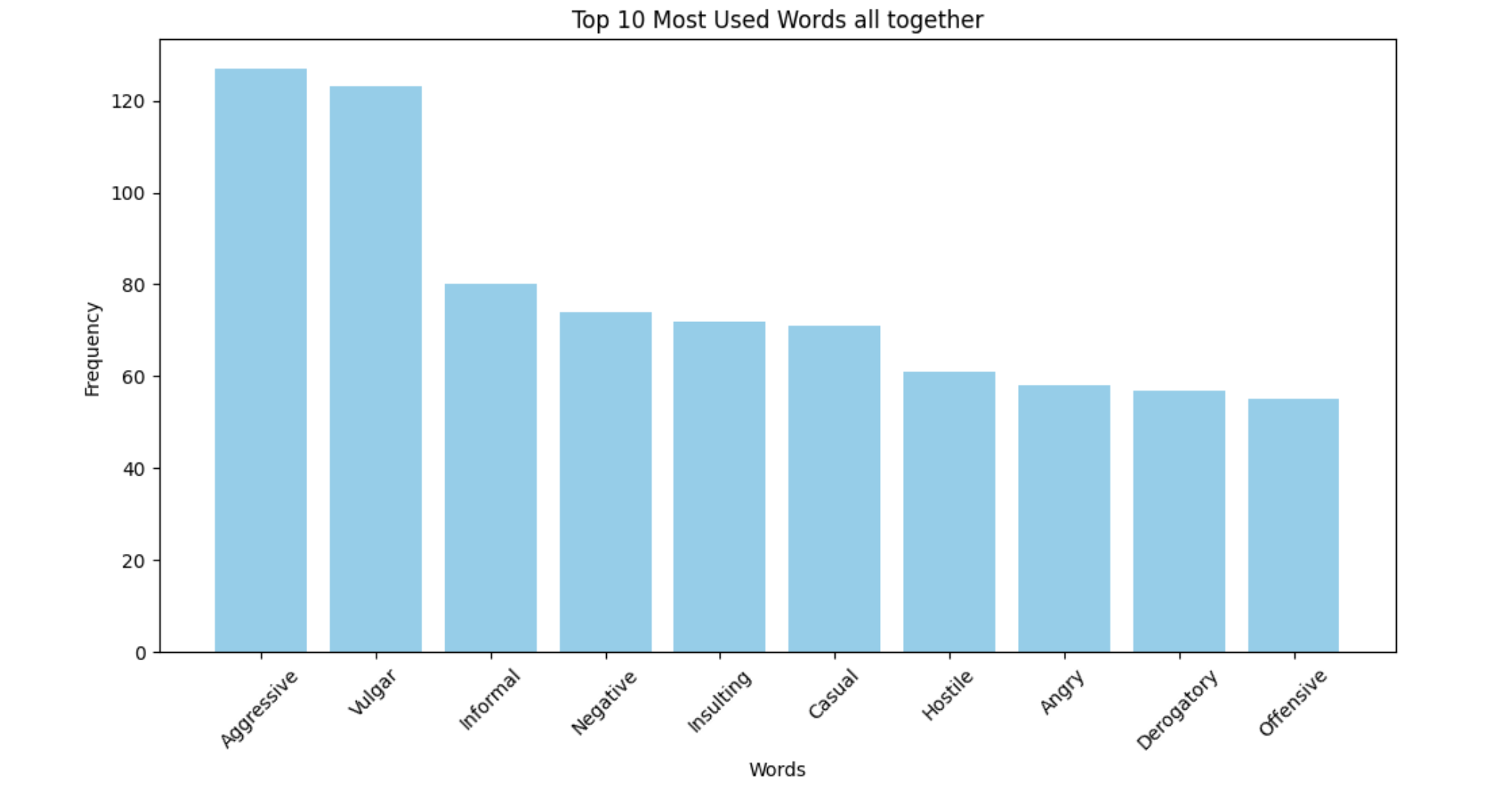}
            \caption[]%
           {{\small All together}}    
            \label{fig:cluster4}
        \end{subfigure}
        \caption[]
       {\small Descriptive words of the different features in the toxic training part for all languages.} 
        \label{fig:train}
    \end{figure*}

    \begin{figure*}
        \centering
        \begin{subfigure}[b]{0.475\textwidth}
            \centering
            \includegraphics[width=\textwidth]{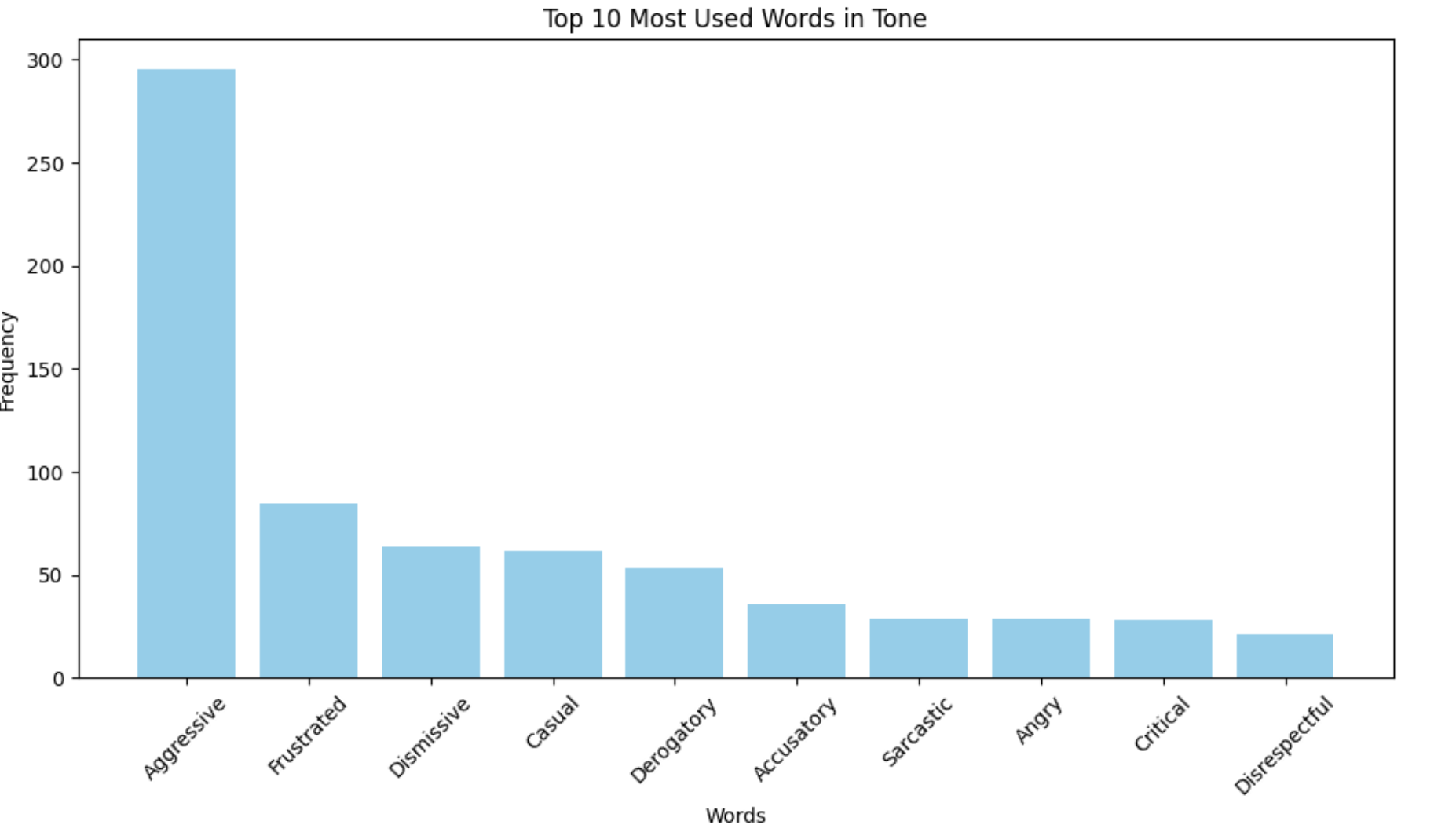}
            \caption[]%
           {{\small Tone}}    
            \label{tone}
        \end{subfigure}
        \hfill
        \begin{subfigure}[b]{0.475\textwidth}  
            \centering 
            \includegraphics[width=\textwidth]{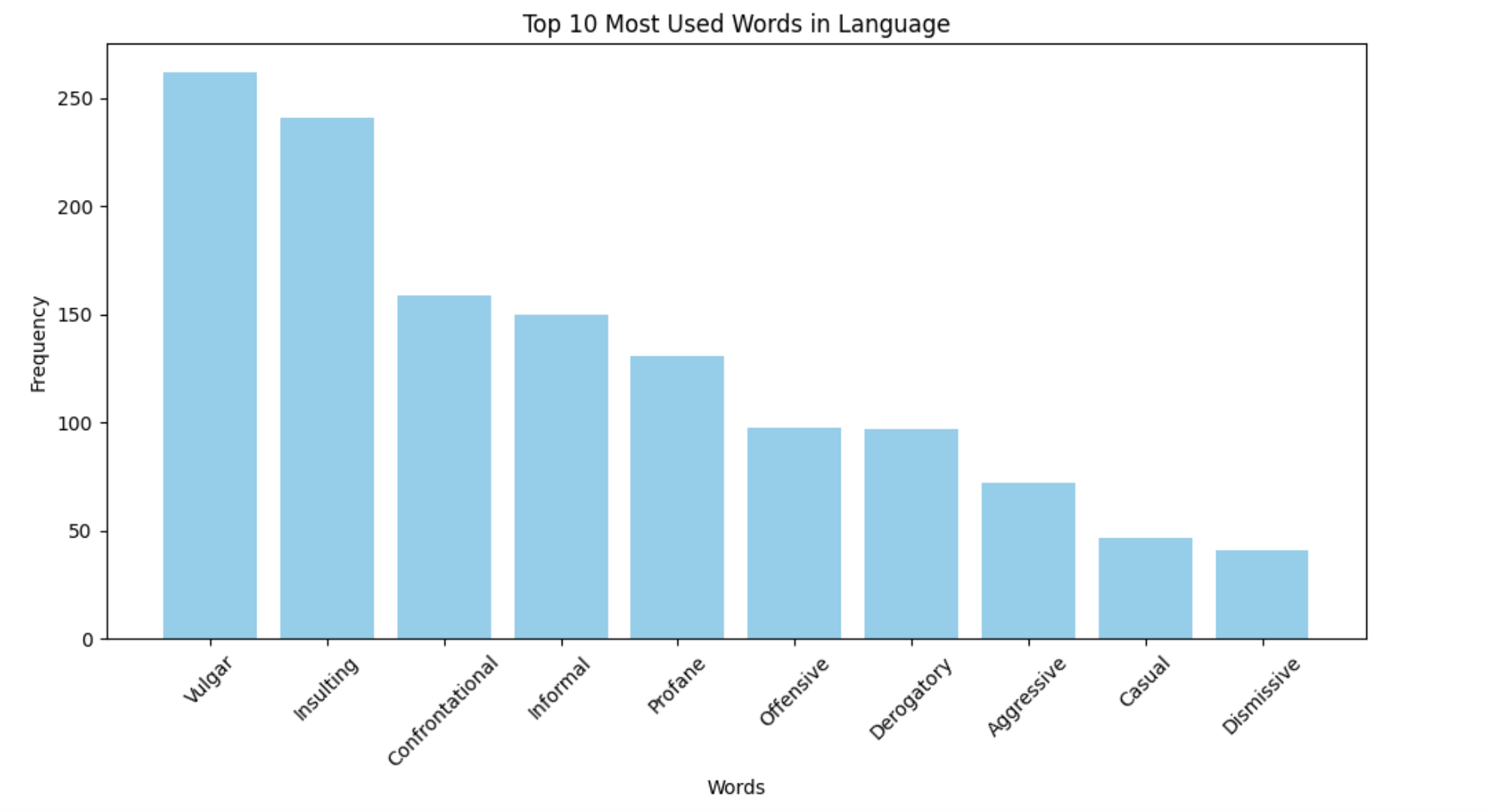}
            \caption[]%
           {{\small Language Type}}    
            \label{fig:cluster2}
        \end{subfigure}
        \vskip\baselineskip
        \begin{subfigure}[b]{0.475\textwidth}   
            \centering 
            \includegraphics[width=\textwidth]{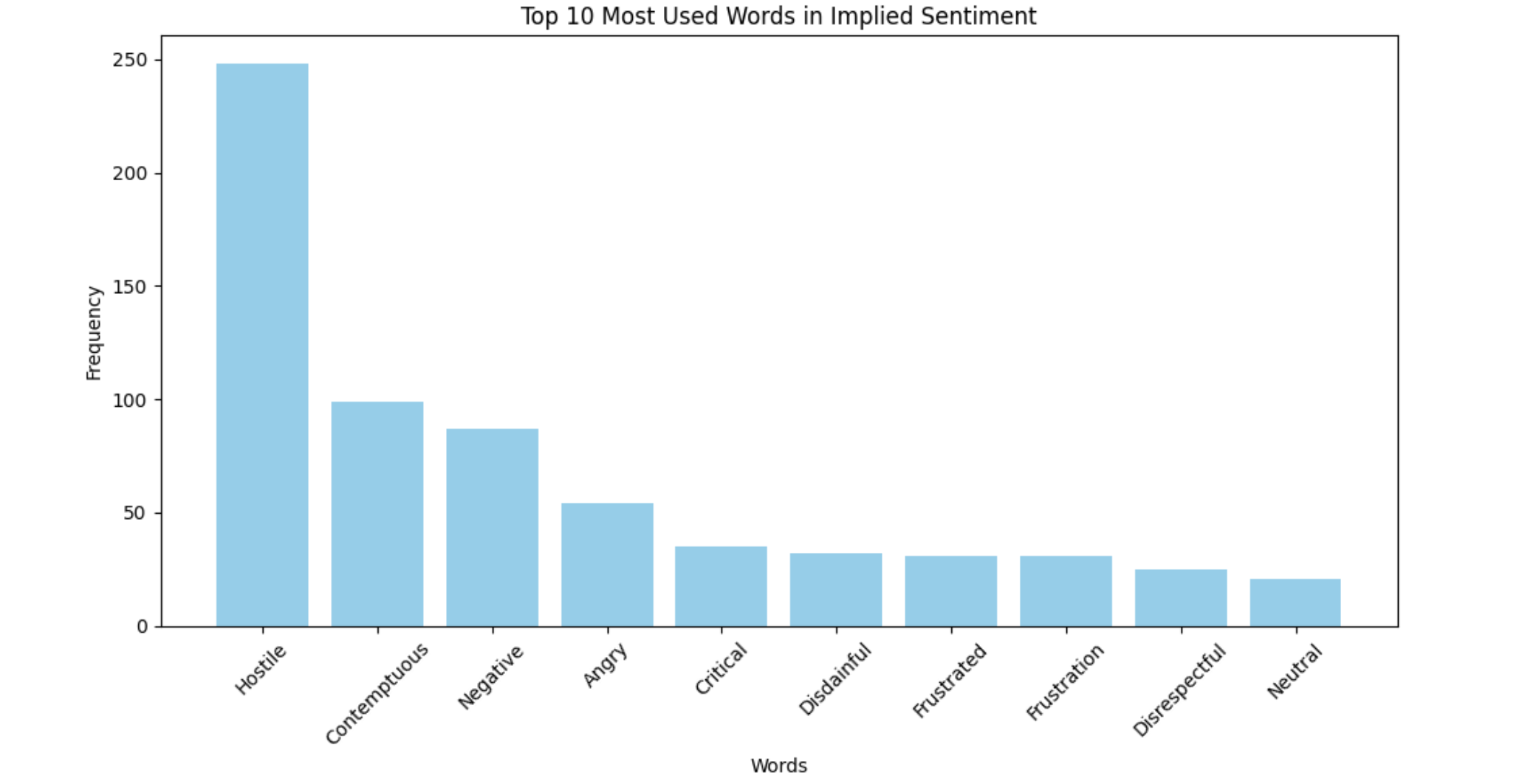}
            \caption[]%
           {{\small Sentiment}}    
            \label{fig:cluster3}
        \end{subfigure}
        \hfill
        \begin{subfigure}[b]{0.475\textwidth}   
            \centering 
            \includegraphics[width=\textwidth]{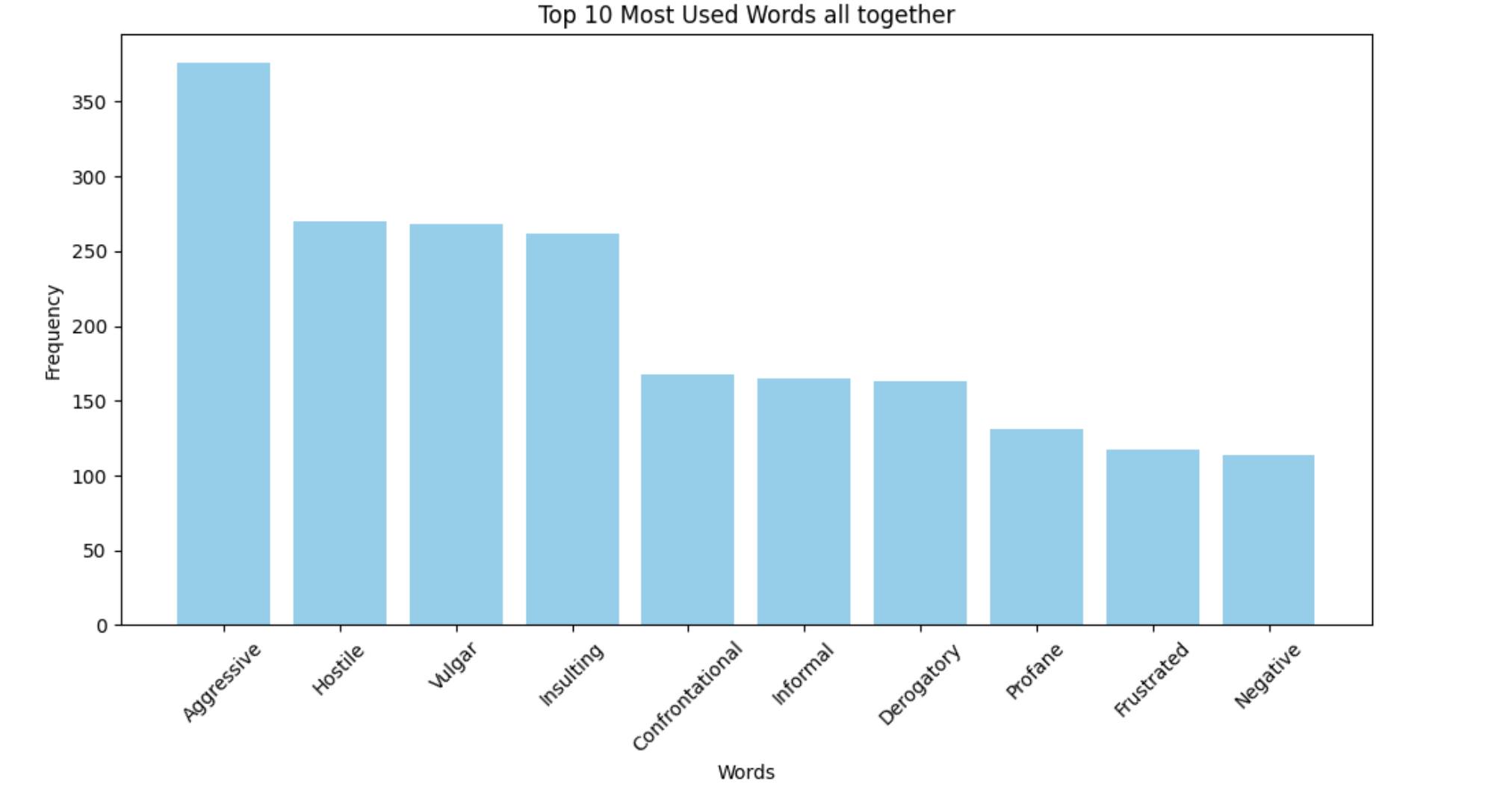}
            \caption[]%
           {{\small All together}}    
            \label{fig:cluster4}
        \end{subfigure}
        \caption[]
       {\small Descriptive words of the different features in the toxic test part for all languages.} 
        \label{fig:test}
    \end{figure*}

\clearpage
\newpage

\section{K-means Clustering Result Examples}
\label{sec:app_clusters}

\begin{figure*}[h!]
    \centering
    \begin{subfigure}[b]{0.475\textwidth}
        \centering
        \includegraphics[width=0.9\textwidth]{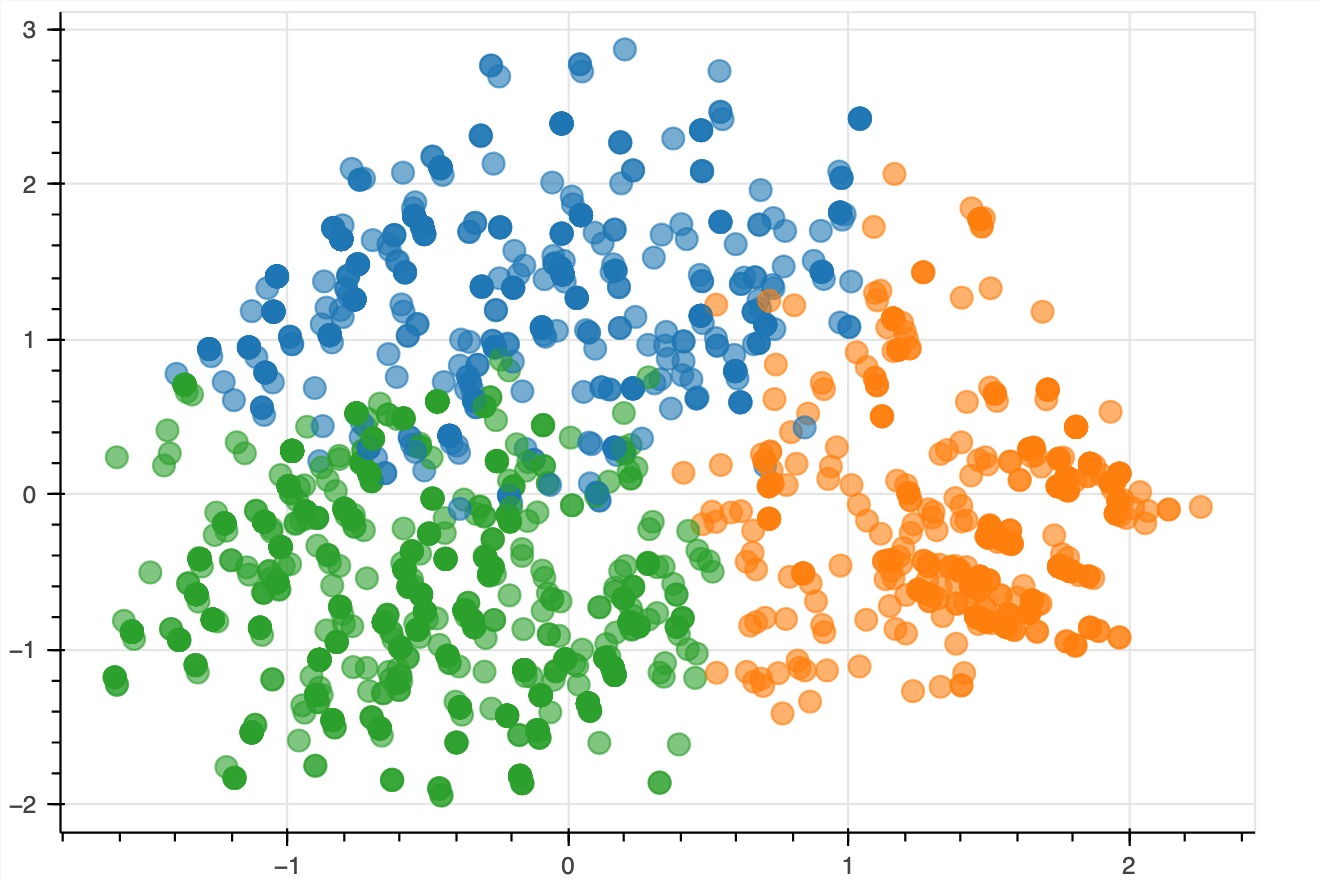}
        \caption[]%
       {{\small All clusters}}    
        \label{fig:cluster1}
    \end{subfigure}
    \hfill
    \begin{subfigure}[b]{0.475\textwidth}  
        \centering 
        \includegraphics[width=\textwidth]{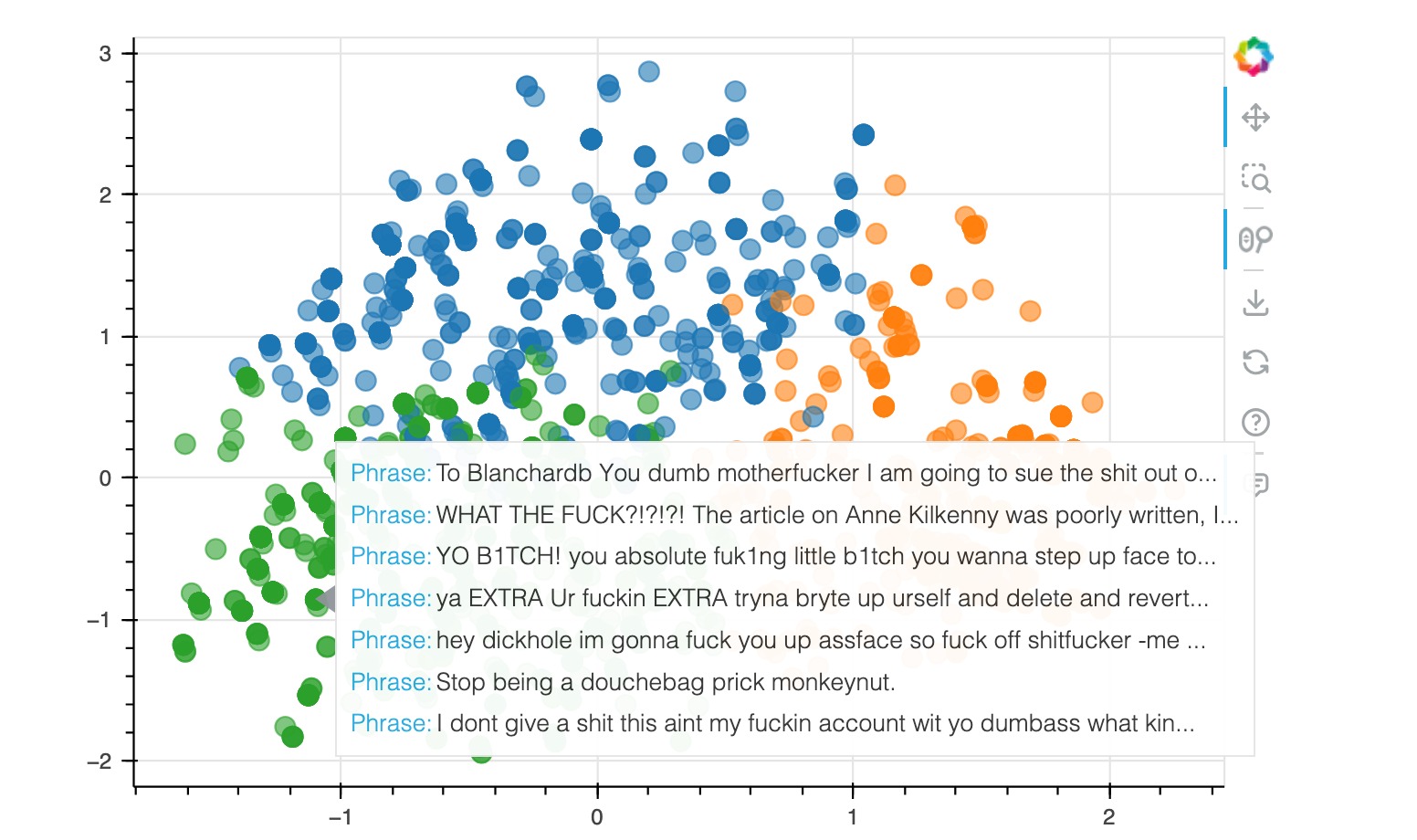}
        \caption[]%
       {{\small Cluster0 examples}}    
        \label{fig:cluster2}
    \end{subfigure}
    \vskip\baselineskip
    \begin{subfigure}[b]{0.475\textwidth}   
        \centering 
        \includegraphics[width=\textwidth]{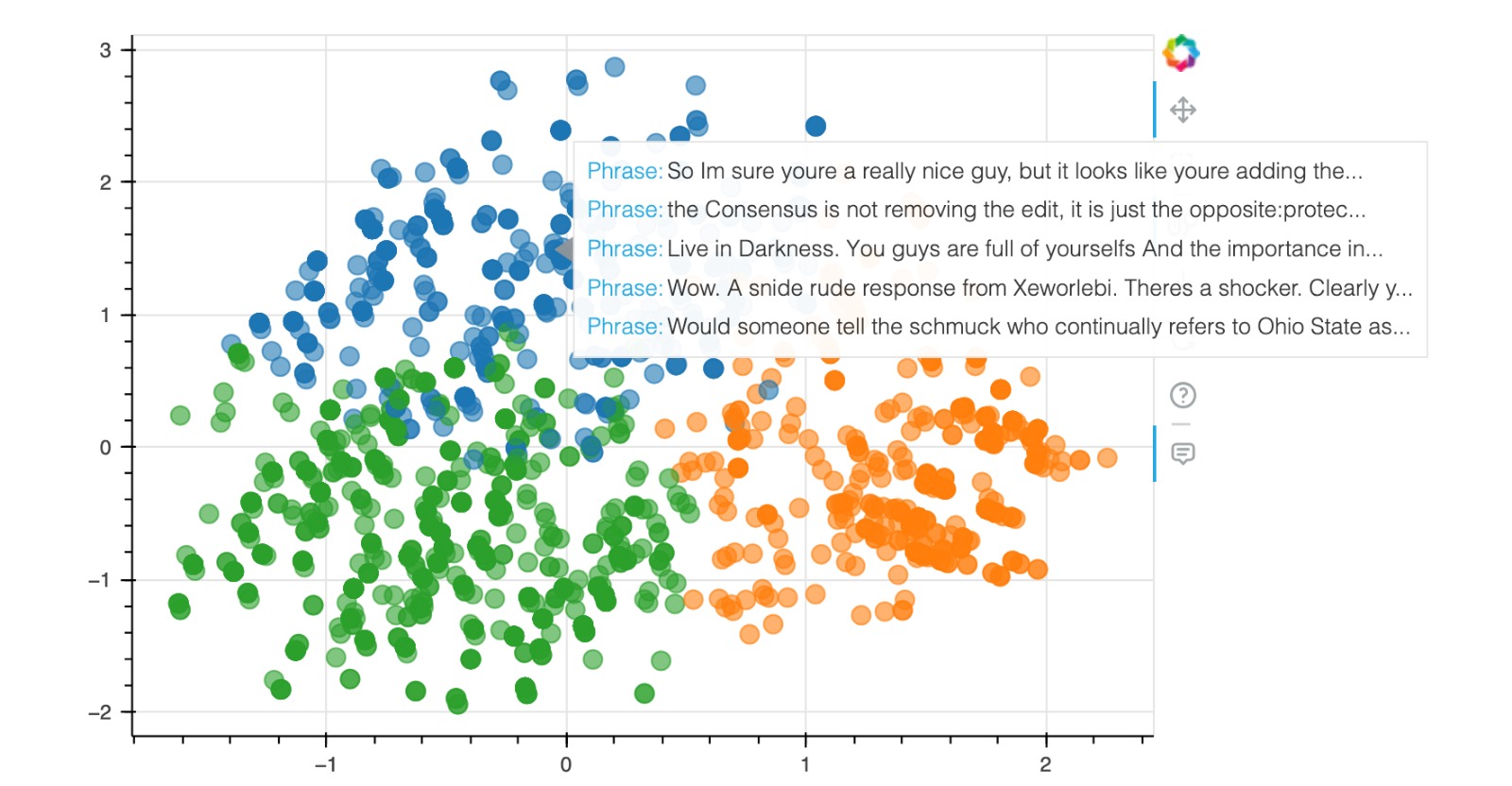}
        \caption[]%
       {{\small Cluster1 examples}}    
        \label{fig:cluster3}
    \end{subfigure}
    \hfill
    \begin{subfigure}[b]{0.475\textwidth}   
        \centering 
        \includegraphics[width=\textwidth]{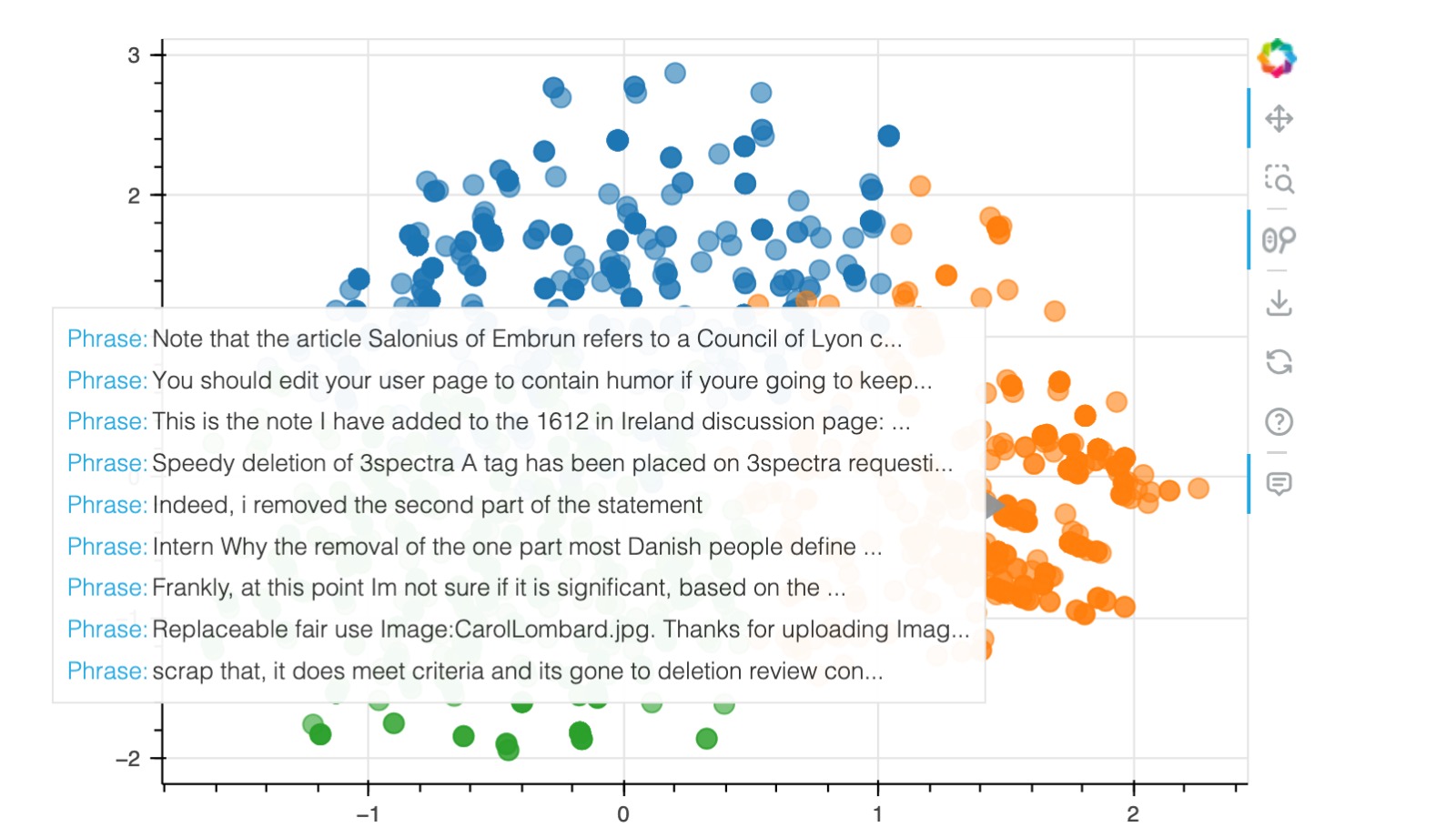}
        \caption[]%
       {{\small Cluster2 examples}}    
        \label{fig:cluster4}
    \end{subfigure}
    \caption[]
   {\small The PCA projection of the toxic sentences cluster based on their descriptive features and detoxification types.} 
    \label{fig:pca}
\end{figure*}

Here, we present the 2D PCA projection of English toxic texts, one-hot-encoded with descriptive features, along with the resulting cluster divisions. (Figure~\ref{fig:pca}).

\section{Automatic Evaluation Results per Language per Metric}
\label{sec:app_results}

Here, we provide the extended results of automatic evaluation setup based on all three evaluation parameters for all languages: English, Spanish, and German (Table~\ref{tab:full_results_english_spanish_german}); Chinese, Arabic, and Hindi (Table~\ref{tab:full_results_chinese_arabic_hindi}); Ukrainian, Russian, and Amharic (Table~\ref{tab:full_results_ukrainian_russian_amharic}).


\begingroup
\renewcommand{\arraystretch}{1.15}
\begin{table}[!ht]
    \centering
    \scriptsize
    \begin{tabular}{l|c c c c|c c c c|c c c c}
    \toprule
    & \multicolumn{4}{c}{\textbf{English}} & \multicolumn{4}{c}{\textbf{Spanish}} & \multicolumn{4}{c}{\textbf{German}} \\
    \midrule
    & \textbf{STA} & \textbf{SIM} & \textbf{ChrF} & \textbf{J} & \textbf{STA} & \textbf{SIM} & \textbf{ChrF} & \textbf{J} & \textbf{STA} & \textbf{SIM} & \textbf{ChrF} & \textbf{J} \\
    \midrule
    \rowcolor{Lightgreen} Human References & $0.864$ & $0.820$ & $1.000$ & $0.711$ & $0.875$ & $0.811$ & $1.000$ & $0.708$ & $0.809$ & $0.909$ & $1.000$ & $0.732$ \\
    \midrule
    & \multicolumn{12}{c}{\textbf{\textit{Unsupervised Approaches}}} \\
    \midrule
    \rowcolor{Gray} Duplicate & $0.090$ & $0.999$ & $0.670$ & $0.061$ & $0.139$ & $0.999$ & $0.655$ & $0.089$ & $0.352$ & $0.999$ & $0.812$ & $0.287$ \\
    Delete & $0.662$ & \underline{$\boldsymbol{0.956}$} & $0.691$ & $0.447$ & $0.479$ & \underline{$\boldsymbol{0.972}$} & \underline{$\boldsymbol{0.669}$} & $0.318$ & $0.454$ & \underline{$\boldsymbol{0.989}$} & \underline{$\boldsymbol{0.802}$} & $\boldsymbol{0.361}$ \\
    Backtranslation & $\boldsymbol{0.807}$ & $0.868$ & $\boldsymbol{0.693}$ & \underline{$\boldsymbol{0.506}$} & $\boldsymbol{0.812}$ & $0.770$ & $0.423$ & $0.275$ & $\boldsymbol{0.796}$ & $0.747$ & $0.372$ & $0.232$ \\
    condBERT & $0.443$ & $0.941$ & $0.640$ & $0.278$ & $0.610$ & $0.920$ & $0.602$ & $\boldsymbol{0.347}$ & $0.419$ & $0.966$ & $0.753$ & $0.310$ \\
    \midrule
    & \multicolumn{12}{c}{\textbf{\textit{Supervised Approaches}}} \\
    \midrule
    mBART-Translated & $\boldsymbol{0.691}$ & $0.894$ & $0.694$ & $\boldsymbol{0.443}$ & $\boldsymbol{0.607}$ & $0.877$ & $0.587$ & $\boldsymbol{0.315}$ & $\boldsymbol{0.581}$ & $0.929$ & $0.729$ & $0.392$ \\
    mBART-mParaDetox & $0.493$ & $\boldsymbol{0.934}$ & \underline{$\boldsymbol{0.695}$} & $0.339$ & $0.474$ & $\boldsymbol{0.933}$ & $\boldsymbol{0.635}$ & $0.289$ & $0.532$ & $\boldsymbol{0.969}$ & $\boldsymbol{0.794}$ & \underline{$\boldsymbol{0.409}$} \\
    \midrule
    & \multicolumn{12}{c}{\textbf{\textit{LLM-based Approaches}}} \\
    \midrule
    GPT-4 few-shot & $0.807$ & $\boldsymbol{0.865}$ & $\boldsymbol{0.661}$ & $\boldsymbol{0.475}$ & $0.867$ & $\boldsymbol{0.806}$ & $\boldsymbol{0.584}$ & $0.421$ & $0.683$ & $\boldsymbol{0.888}$ & $\boldsymbol{0.659}$ & $0.395$ \\
    GPT-4 CoT &  \underline{$\boldsymbol{0.985}$} & $0.682$ & $0.454$ & $0.326$ & \underline{$\boldsymbol{0.949}$} & $0.789$ & $0.573$ & \underline{$\boldsymbol{0.447}$} & \underline{$\boldsymbol{0.908}$} & $0.783$ & $0.544$ & $\boldsymbol{0.400}$ \\
    \bottomrule
    \end{tabular}
    \caption{Automatic evaluation results for English, Spanish, and German. \textbf{Bold} denote the best results within the group, \underline{\textbf{underlined}}---the best for the language.}
    \label{tab:full_results_english_spanish_german}
\end{table}
\endgroup

\begingroup
\renewcommand{\arraystretch}{1.15}
\begin{table}[!ht]
    \centering
    \scriptsize
    \begin{tabular}{l|c c c c|c c c c|c c c c}
    \toprule
    & \multicolumn{4}{c}{\textbf{Chinese}} & \multicolumn{4}{c}{\textbf{Arabic}} & \multicolumn{4}{c}{\textbf{Hindi}} \\
    \midrule
    & \textbf{STA} & \textbf{SIM} & \textbf{ChrF} & \textbf{J} & \textbf{STA} & \textbf{SIM} & \textbf{ChrF} & \textbf{J} & \textbf{STA} & \textbf{SIM} & \textbf{ChrF} & \textbf{J} \\
    \midrule
    \rowcolor{Lightgreen} Human References & $0.266$ & $0.789$ & $1.000$ & $0.201$ & $0.795$ & $0.875$ & $1.000$ & $0.694$ & $0.367$ & $0.814$ & $1.000$ & $0.297$ \\
    \midrule
    & \multicolumn{12}{c}{\textbf{\textit{Unsupervised Approaches}}} \\
    \midrule
    \rowcolor{Gray} Duplicate & $0.130$ & $0.999$ & $0.535$ & $0.069$ & $0.388$ & $0.999$ & $0.776$ & $0.293$ & $0.051$ & $0.999$ & $0.695$ & $0.034$ \\
    Delete & $0.384$ & $0.887$ & \underline{$\boldsymbol{0.524}$} & \underline{$\boldsymbol{0.174}$} & $0.597$ & \underline{$\boldsymbol{0.974}$} & \underline{$\boldsymbol{0.777}$} & \underline{$\boldsymbol{0.455}$} & $0.146$ & $0.974$ & \underline{$\boldsymbol{0.706}$} & $\boldsymbol{0.104}$\\
    Backtranslation & $\boldsymbol{0.661}$ & $0.591$ & $0.070$ & $0.026$ & $\boldsymbol{0.836}$ & $0.682$ & $0.319$ & $0.205$ & $\boldsymbol{0.443}$ & $0.731$ & $0.289$ & $0.103$ \\
    condBERT        & $0.138$ & \underline{$\boldsymbol{0.993}$} & $0.518$ & $0.067$ & $0.488$ & $0.957$ & $0.726$ & $0.337$ & $0.050$ & \underline{$\boldsymbol{0.976}$} & $0.667$ & $0.033$ \\
    \midrule
    & \multicolumn{12}{c}{\textbf{\textit{Supervised Approaches}}} \\
    \midrule
    mBART-Translated & $\boldsymbol{0.272}$ & $0.901$ & $0.356$ & $\boldsymbol{0.083}$ & $\boldsymbol{0.626}$ & $0.899$ & $0.667$ & $0.365$ & $\boldsymbol{0.243}$ & $0.896$ & $0.617$ & $0.142$ \\
    mBART-mParaDetox & $0.166$ & $\boldsymbol{0.963}$ & $\boldsymbol{0.433}$ & $0.068$ & $0.560$ & $\boldsymbol{0.950}$ & $\boldsymbol{0.742}$ & $\boldsymbol{0.397}$ & $0.234$ & $\boldsymbol{0.939}$ & $\boldsymbol{0.699}$ & $\boldsymbol{0.171}$ \\
    \midrule
    & \multicolumn{12}{c}{\textbf{\textit{LLM-based Approaches}}} \\
    \midrule
    GPT-4 few-shot & $0.452$ & $\boldsymbol{0.805}$ & $\boldsymbol{0.328}$ & $0.108$ & $0.759$ & $\boldsymbol{0.755}$ & $0.466$ & $0.270$ & $0.476$ & $\boldsymbol{0.786}$ & $0.509$ & $0.193$ \\
    GPT-4 CoT & \underline{$\boldsymbol{0.716}$} & $0.683$ & $0.228$ & $\boldsymbol{0.117}$ & \underline{$\boldsymbol{0.931}$} & $0.712$ & $\boldsymbol{0.476}$ & $\boldsymbol{0.339}$ & \underline{$\boldsymbol{0.611}$} & $0.745$ & $\boldsymbol{0.533}$ & \underline{$\boldsymbol{0.251}$} \\
    \bottomrule
    \end{tabular}
    \caption{Automatic evaluation results for Chinese, Arabic, and Hindi. \textbf{Bold} denote the best results within the group, \underline{\textbf{underlined}}---the best for the language.}
    \label{tab:full_results_chinese_arabic_hindi}
\end{table}
\endgroup

\begingroup
\renewcommand{\arraystretch}{1.15}
\begin{table}[!ht]
    \centering
    \scriptsize
    \begin{tabular}{l|c c c c| c c c c|c c c c}
    \toprule
    & \multicolumn{4}{c}{\textbf{Ukrainian}} & \multicolumn{4}{c}{\textbf{Russian}} & \multicolumn{4}{c}{\textbf{Amharic}} \\
    \midrule
    & \textbf{STA} & \textbf{SIM} & \textbf{ChrF} & \textbf{J} & \textbf{STA} & \textbf{SIM} & \textbf{ChrF} & \textbf{J} & \textbf{STA} & \textbf{SIM} & \textbf{ChrF} & \textbf{J} \\
    \midrule
    \rowcolor{Lightgreen} Human References & $0.877$ & $0.899$ & $1.000$ & $0.790$ & $0.887$ & $0.824$ & $1.000$ & $0.732$ & $0.893$ & $0.683$ & $1.000$ & $0.601$ \\
    \midrule
    & \multicolumn{12}{c}{\textbf{\textit{Unsupervised Approaches}}} \\
    \midrule
    \rowcolor{Gray} Duplicate & $0.037$ & $0.999$ & $0.778$ & $0.031$ & $0.067$ & $0.999$ & $0.698$ & $0.048$ & $0.426$ & $0.999$ & $0.485$ & $0.216$\\
    Delete & $0.423$ & \underline{$\boldsymbol{0.974}$} & \underline{$\boldsymbol{0.791}$} & $\boldsymbol{0.327}$ & $0.372$ & \underline{$\boldsymbol{0.971}$} & $\boldsymbol{0.708}$ & $\boldsymbol{0.254}$ & $0.539$ & \underline{$\boldsymbol{0.979}$} & \underline{$\boldsymbol{0.486}$} & \underline{$\boldsymbol{0.269}$} \\
    Backtranslation & $\boldsymbol{0.914}$ & $0.704$ & $0.293$ & $0.201$ & $\boldsymbol{0.903}$ & $0.697$ & $0.328$ & $0.222$ & $0.819$ & $0.618$ & $0.135$ & $0.075$ \\
    condBERT         & $0.424$ & $0.960$ & $0.759$ & $0.316$ & $0.339$ & $0.944$ & $0.666$ & $0.224$ & \underline{$\boldsymbol{0.998}$} & $0.169$ & $0.007$ & $0.003$ \\
    \midrule
    & \multicolumn{12}{c}{\textbf{\textit{Supervised Approaches}}} \\
    \midrule
    mBART-Translated & $\boldsymbol{0.610}$ & $0.870$ & $0.647$ & $0.343$ & $\boldsymbol{0.601}$ & $0.885$ & $0.657$ & $\boldsymbol{0.359}$ & $0.501$ & $\boldsymbol{0.875}$ & $0.391$ & $0.178$ \\
    mBART-mParaDetox & $0.462$ & $\boldsymbol{0.939}$ & $\boldsymbol{0.751}$ & $\boldsymbol{0.345}$ & $0.455$ & $\boldsymbol{0.937}$ & \underline{$\boldsymbol{0.709}$} & $0.321$ & $\boldsymbol{0.506}$ & $0.915$ & $\boldsymbol{0.412}$ & $\boldsymbol{0.204}$ \\
    \midrule
    & \multicolumn{12}{c}{\textbf{\textit{LLM-based Approaches}}} \\
    \midrule
    GPT-4 few-shot & $0.895$ & $\boldsymbol{0.842}$ & $0.592$ & $0.460$ & $0.758$ & $\boldsymbol{0.825}$ & $\boldsymbol{0.600}$ & $0.382$ & $0.467$ & $\boldsymbol{0.946}$ & $\boldsymbol{0.453}$ & $\boldsymbol{0.205}$ \\
    GPT-4 CoT & \underline{$\boldsymbol{0.955}$} & $0.823$ & $\boldsymbol{0.612}$ & \underline{$\boldsymbol{0.503}$} & \underline{$\boldsymbol{0.939}$} & $0.782$ & $0.560$ & \underline{$\boldsymbol{0.426}$} & $\boldsymbol{0.835}$ & $0.672$ & $0.269$ & $0.166$ \\
    \bottomrule
    \end{tabular}
    \caption{Automatic evaluation results for Ukrainian, Russian, and Amharic. \textbf{Bold} denote the best results within the group, \underline{\textbf{underlined}}---the best for the language.}
    \label{tab:full_results_ukrainian_russian_amharic}
\end{table}
\endgroup

\begin{table*}[h!]
    \footnotesize
    \centering
    \begin{tabular}{p{3.5cm}|p{10cm}}
        \toprule 
        \rowcolor{Gray} Original &  and nobody gave a sh*t .\\
        \midrule
        Delete & and nobody gave a . \\
        BART-detox (en) & and nobody cared. \\
        CondBERT & and nobody gave a buttu . \\
        mBART-Translated & and nobody gave a d*mn. \\
        mBART-ParaDetox & and nobody gave a sh*t. \\
        GPT-4 few-shot & and nobody cared. \\
        GPT-4 CoT & and nobody seemed to care. \\
        \midrule

        \rowcolor{Gray} Original & everyone s got that single issue that represents the burr up their a*s  \\
        \midrule
        Delete & everyone s got that single issue that represents the burr up their . \\
        BART-detox (en) & everyone s got that single issue that represent the burr up. \\
        CondBERT & everyone s got that single issue that represents the burr up their Ass . \\
        mBART-Translated & everyone s got that single issue that represents the burr up their a*s \\
        mBART-ParaDetox & everyone s got that single issue that represents the burr up their a*s. \\
        GPT-4 few-shot & everyone has that one issue that bothers them. \\
        GPT-4 CoT & Everyone has that one issue that really bothers them. \\

        \bottomrule
    \end{tabular}
    \caption{Examples of text detoxification outputs by different models for English for general readers to showcase the approached behaviour. For the phrases that require significant rephrasing, LLM, especially, with proposed CoT method suggests more reasonable detoxification. For mBART, it seems challenging to grasp detoxification knowledge properly for nine languages simultaneously.}
    \label{tab:output_examples_english}
\end{table*}

\newpage

\section{Multilingual ParaDetox Data Examples}
\label{sec:app_data_examples}

\begin{figure}[h!]
    \centering
    \includegraphics[width=0.875\textwidth]{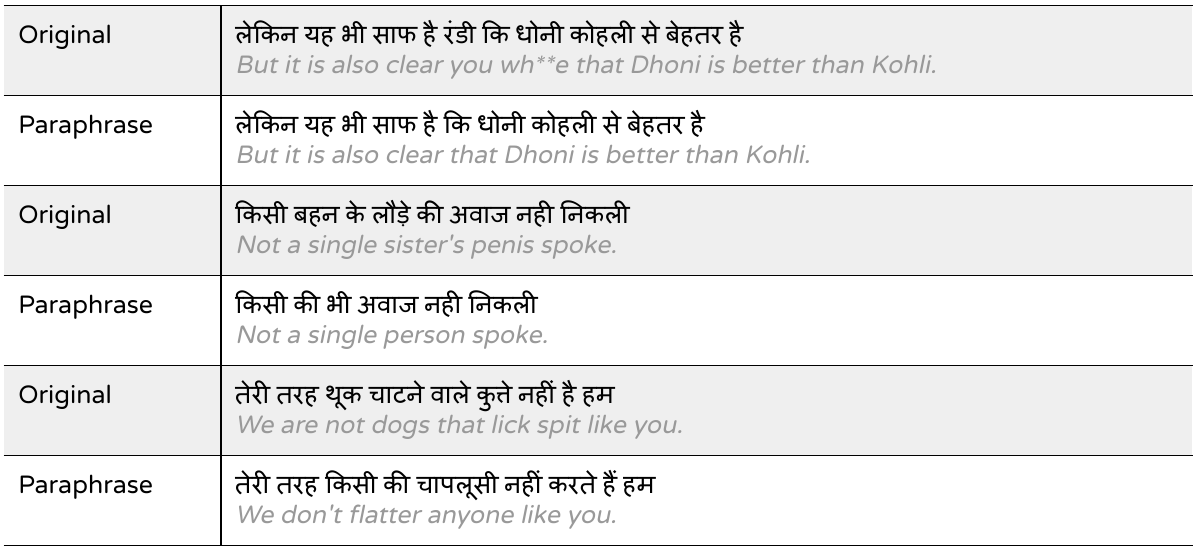}
    \caption{Examples of parallel detoxified pairs from HiParaDetox.}
    \label{fig:parallel_examples_hindi}
\end{figure}
\begin{figure}[h!]
    \centering
    \includegraphics[width=0.875\textwidth]{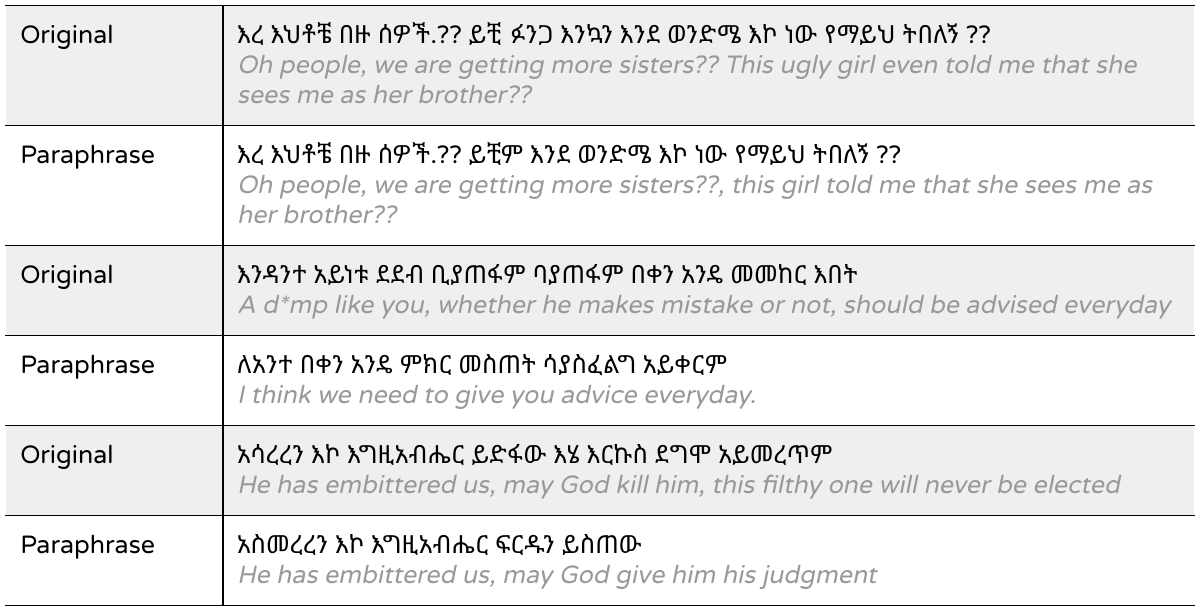}
    \caption{Examples of parallel detoxified pairs from AmParaDetox.}
    \label{fig:parallel_examples_amharic}
\end{figure}
\begin{figure}[h!]
    \centering
    \includegraphics[width=0.875\textwidth]{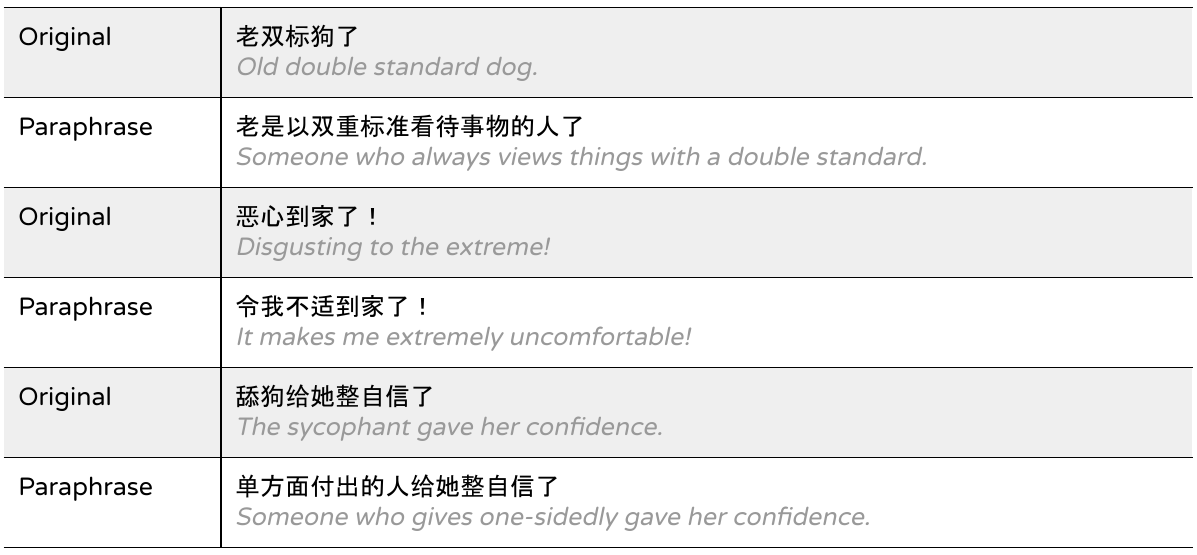}
    \caption{Examples of parallel detoxified pairs from ZhParaDetox.}
    \label{fig:parallel_examples_chinese}
\end{figure}
\begin{figure}[h!]
    \centering
    \includegraphics[width=0.85\textwidth]{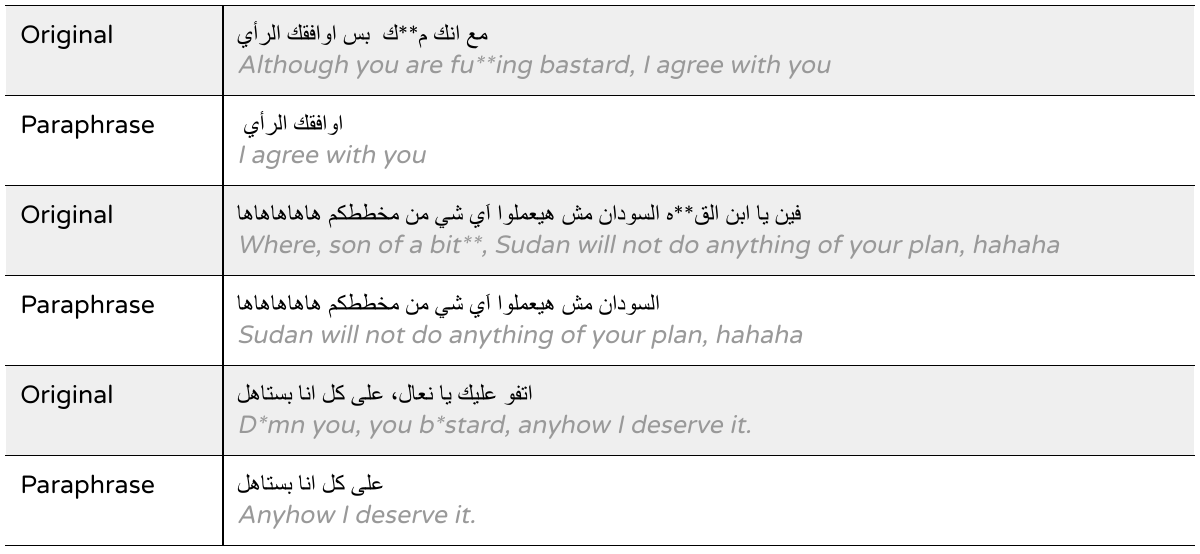}
    \caption{Examples of parallel detoxified pairs from ArParaDetox.}
    \label{fig:parallel_examples_arabic}
\end{figure}

Here, we provide an example with extracted features for English (Table~\ref{tab:parallel_examples_english}) for general readers and several examples of data samples from new collected parallel text detoxification data for new languages: German (Table~\ref{tab:parallel_examples_german}), Hindi (Table~\ref{fig:parallel_examples_hindi}), Amharic (Table~\ref{fig:parallel_examples_amharic}), Chinese (Table~\ref{fig:parallel_examples_chinese}), and Arabic (Table~\ref{fig:parallel_examples_arabic}).

\begin{table*}[h!]
    \centering
    \begin{tabular}{p{3cm}|p{10cm}}
        \toprule 
        \rowcolor{Gray} Original & You made a mistake you a**. \\
        \midrule
        \textit{Toxicity Level} & High \\
        \textit{Tone} & Aggressive \\
        \textit{Language} & Insulting and offensive \\
        \textit{Implied Sentiment} & Hostile \\
        \textit{Toxic Keywords} & a** \\ 
        \midrule
        \rowcolor{Lightgreen} Paraphrase & You made a mistake. \\
        \midrule
        \textit{Toxicity Level} & Low \\
        \textit{Tone} & Assertive \\
        \textit{Language} & Informal and straightforward \\
        \textit{Implied Sentiment} & Negative \\
        \textit{Toxic Keywords} & None \\ 
        \midrule

        \rowcolor{Gray} Original & holy sh*t i thought that was a joke . \\
        \midrule
        \textit{Toxicity Level} & Medium \\
        \textit{Tone} & Casual \\
        \textit{Language} & Informal \\
        \textit{Implied Sentiment} & Surprised \\
        \textit{Toxic Keywords} & sh*t \\ 
        \midrule
        \rowcolor{Lightgreen} Paraphrase & I thought it was a joke!! \\
        \midrule
        \textit{Toxicity Level} & Low \\
        \textit{Tone} & Casual \\
        \textit{Language} & Informal \\
        \textit{Implied Sentiment} & Surprise or skepticism \\
        \textit{Toxic Keywords} & None \\ 
        
        
        \bottomrule
    \end{tabular}
    \caption{Examples of parallel detoxified pairs from EnParaDetox.}
    \label{tab:parallel_examples_english}
\end{table*}

        


\begin{table*}[ht!]
    \centering
    \begin{tabular}{p{2cm}|p{11cm}}
        \toprule 
        \rowcolor{Gray} Original & \foreignlanguage{german}{idi*****her Kommentar. Aufm Supermannheft gepennt?} \newline \textcolor{gray}{\scriptsize{\textit{Idi***c comment. Slipped up on the Superman magazine?}}}\\
        \midrule
        Paraphrase & \foreignlanguage{german}{schlechter Kommentar. Aufm Supermannheft gepennt?} \newline \textcolor{gray}{\scriptsize{\textit{bad comment. Slipped up on the Superman magazine?}}} \\
        \midrule
        \rowcolor{Gray} Original & \foreignlanguage{german}{Ich will dieses A*****och nicht auf freiem Fuß wissen...egal in welchem Land..!} \newline \textcolor{gray}{\scriptsize{\textit{I don't want this a***ole at liberty...no matter in which country...!}}}\\
        \midrule
        Paraphrase & \foreignlanguage{german}{Ich will diese Person nicht auf freiem Fuß wissen...egal in welchem Land..!} \newline \textcolor{gray}{\scriptsize{\textit{I don't want this person at liberty...no matter in which country...!}}} \\
        \midrule

        \rowcolor{Gray} Original & \foreignlanguage{german}{Ich finde er ist einfach ein unlustiger Spas**i} \newline \textcolor{gray}{\scriptsize{\textit{I just think he's an unfunny ret**d}}}\\
        \midrule
        Paraphrase & \foreignlanguage{german}{Ich finde er ist einfach nicht lustig} \newline \textcolor{gray}{\scriptsize{\textit{I just don't think he's funny}}} \\
        \bottomrule
    \end{tabular}
    \caption{Examples of parallel detoxified pairs from DeParaDetox.}
    \label{tab:parallel_examples_german}
\end{table*}

\clearpage

\end{document}